\documentclass[acmsmall]{acmart}
\usepackage{amsmath}

\usepackage{amssymb}
\usepackage{booktabs}
\usepackage{multirow}
\usepackage{algorithm}
\usepackage{algorithmic}
\usepackage{setspace}
\usepackage{float}
\usepackage{geometry}
\usepackage{graphicx}
\usepackage{subcaption}
\newtheorem{theorem}{\textbf{Theorem}}
\newtheorem{definition}{\textbf{Definition}}
\newtheorem{lemma}{\textbf{Lemma}}
\newtheorem{example}{\textbf{Example}}
\newtheorem{assumption}{\textbf{Assumption}}

\AtBeginDocument{%
  }

\setcopyright{acmcopyright}
\copyrightyear{2018}
\acmYear{2018}
\acmDOI{XXXXXXX.XXXXXXX}

\acmJournal{JACM}
\acmVolume{37}
\acmNumber{4}
\acmArticle{111}
\acmMonth{8}

\begin{document}

%%
%% The "title" command has an optional parameter,
%% allowing the author to define a "short title" to be used in page headers.
\title{Fair Causal Feature Selection}

%%
%% The "author" command and its associated commands are used to define
%% the authors and their affiliations.
%% Of note is the shared affiliation of the first two authors, and the
%% "authornote" and "authornotemark" commands
%% used to denote shared contribution to the research.
\author{Zhaolong~Ling}
%\authornote{Both authors contributed equally to this research.}

%\orcid{1234-5678-9012}
\author{Enqi~Xu}

\author{Peng~Zhou}
\authornote{This is the corresponding author}
\affiliation{%
  \institution{Anhui University}
  %\streetaddress{P.O. Box 1212}
  \city{Hefei}
  %\state{Ohio}
  \country{China}
  \postcode{230601}
}
\email{zlling@ahu.edu.cn}
\email{enqixu@ahu.edu.cn}
\email{zhoupeng@ahu.edu.cn}

\author{Liang~Du}
\affiliation{%
  \institution{School of Computer and Information Technology, Shanxi University}
  %\streetaddress{P.O. Box 1212}
  \city{Taiyuan}
  %\state{Ohio}
  \country{China}
  \postcode{030006}
}
\email{duliang@sxu.edu.cn}
\author{Kui~Yu}

\author{Xindong~Wu}

\affiliation{%
\city{Key Laboratory of Knowledge Engineering  with Big Data (the Ministry of Education of China), and the School of Computer Science and Information Technology}
 \institution{Hefei University of Technology}
  %\streetaddress{111 Jiulong Road}
  \city{Hefei}
  \country{China}}
  \postcode{230009}
\email{yukui@hfut.edu.cn}
\email{wuxindong@mininglamp.com}
%\author{Valerie B\'eranger}
%\affiliation{%
%  \institution{Inria Paris-Rocquencourt}
%  \city{Rocquencourt}
%  \country{France}
%}
%
%\author{Aparna Patel}
%\affiliation{%
% \institution{Rajiv Gandhi University}
% \streetaddress{Rono-Hills}
% \city{Doimukh}
% \state{Arunachal Pradesh}
% \country{India}}
%
%\author{Huifen Chan}
%\affiliation{%
%  \institution{Tsinghua University}
%  \streetaddress{30 Shuangqing Rd}
%  \city{Haidian Qu}
%  \state{Beijing Shi}
%  \country{China}}
%
%\author{Charles Palmer}
%\affiliation{%
%  \institution{Palmer Research Laboratories}
%  \streetaddress{8600 Datapoint Drive}
%  \city{San Antonio}
%  \state{Texas}
%  \country{USA}
%  \postcode{78229}}
%\email{cpalmer@prl.com}
%
%\author{John Smith}
%\affiliation{%
%  \institution{The Th{\o}rv{\"a}ld Group}
%  \streetaddress{1 Th{\o}rv{\"a}ld Circle}
%  \city{Hekla}
%  \country{Iceland}}
%\email{jsmith@affiliation.org}
%
%\author{Julius P. Kumquat}
%\affiliation{%
%  \institution{The Kumquat Consortium}
%  \city{New York}
%  \country{USA}}
%\email{jpkumquat@consortium.net}

%%
%% By default, the full list of authors will be used in the page
%% headers. Often, this list is too long, and will overlap
%% other information printed in the page headers. This command allows
%% the author to define a more concise list
%% of authors' names for this purpose.
\renewcommand{\shortauthors}{Z. Ling et al.}

%%
%% The abstract is a short summary of the work to be presented in the
%% article.
\begin{abstract}
  Fair feature selection for classification decision tasks has recently garnered significant attention from researchers. However, existing fair feature selection algorithms fall short of providing a full explanation of the causal relationship between features and sensitive attributes, potentially impacting the accuracy of fair feature identification. To address this issue, we propose a Fair Causal Feature Selection algorithm, called FairCFS. Specifically, FairCFS constructs a localized causal graph that identifies the Markov blankets of class and sensitive variables, to block the transmission of sensitive information for selecting fair causal features. Extensive experiments on seven public real-world datasets validate that FairCFS has comparable accuracy compared to eight state-of-the-art feature selection algorithms, while presenting more superior fairness.
\end{abstract}

%%
%% The code below is generated by the tool at http://dl.acm.org/ccs.cfm.
%% Please copy and paste the code instead of the example below.
%%
\begin{CCSXML}
<ccs2012>
   <concept>
       <concept_id>10010147.10010178.10010187.10010192</concept_id>
       <concept_desc>Computing methodologies~Causal reasoning and diagnostics</concept_desc>
       <concept_significance>500</concept_significance>
       </concept>
 </ccs2012>
\end{CCSXML}

\ccsdesc[500]{Computing methodologies~Causal reasoning and diagnostics}

%%
%% Keywords. The author(s) should pick words that accurately describe
%% the work being presented. Separate the keywords with commas.
\keywords{Causal Fairness, Fair Feature Selection, Markov Blanket}

\received{20 February 2007}
\received[revised]{12 March 2009}
\received[accepted]{5 June 2009}

%%
%% This command processes the author and affiliation and title
%% information and builds the first part of the formatted document.
\maketitle

\section{Introduction}
Fairness in trustworthy decision-making has become a critical focus~\cite{corbett2017algorithmic}. It ensures that models refrain from discriminating against or demonstrating unfair preferences towards specific individuals or groups grounded in sensitive attributes such as race, gender, or age in various fields such as credit assessment~\cite{bernanke1988credit}, judicature~\cite{sutera1993history}, medicine~\cite{pellegrino1966medicine}. People can now benefit from the convenience of decision-support algorithms and do not worry about the discrimination of specific groups and public awareness~\cite{mehrabi2021survey}. However, existing fairness algorithms aim to mitigate unfair biases without considering the issue of scalability as the dataset size increases~\cite{corbett2018measure}. Thus, researchers have turned to fair feature selection algorithms to address discrimination in fair decision-making tasks~\cite{belitz2021automating,galhotra2022causal}, and identifying a subset of fair features from the original feature set offers a more convenient model-building and data-understanding approach~\cite{grgic2016case}.

Among the existing fair feature selection algorithms, one method involves incorporating fairness metrics into the training process to select suitable features, leveraging traditional feature selection algorithms~\cite{belitz2021automating}. This method adopts a forward feature selection strategy based on the conventional wrapper feature selection method~\cite{kusner2017counterfactual}, and utilizes a combined metric of accuracy and fairness during the model evaluation stage for selecting fair features. Although this method of selecting features based on statistical metrics demonstrates their utility in enhancing both the accuracy and fairness of the predictive model, it can not explain the causal relationship between these features and sensitive variables.

\begin{figure}[t]
  \centering
  \includegraphics[height=2.0in]{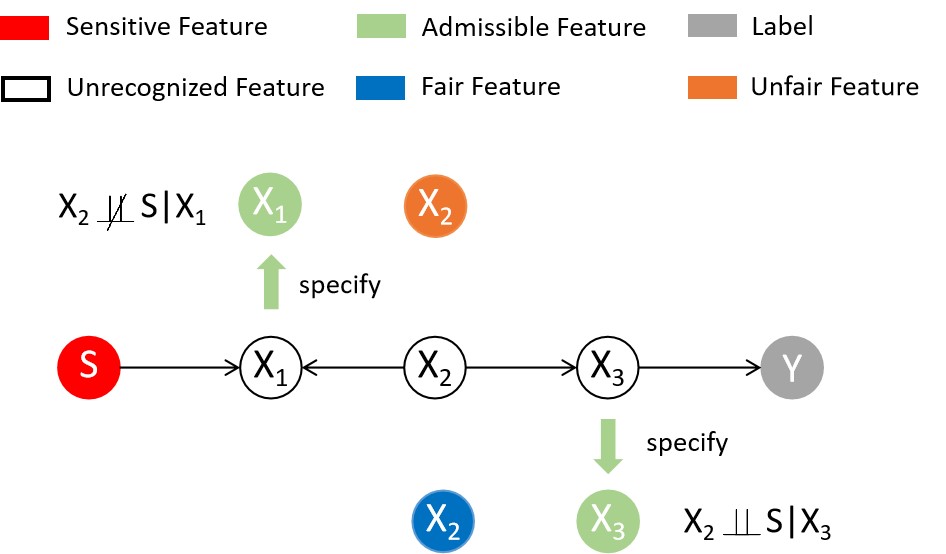}\\
   \caption{\textbf{ The difference in admissible features leads to different judgments on whether a feature is fair or not.} (a) When $X_{1}$ is specified as an admissible feature, $X_{2}$ is dependent on the sensitive variable $S$, and then $X_{2}$ is an unfair feature. (b) But when $X_{3}$ is specified as an acceptable feature, $X_{2}$ is independent of the sensitive variable $S$, and then $X_{2}$ becomes a fair feature.}
\end{figure}

Other researchers have proposed a fairness feature selection method that adheres to the definition of causal fairness~\cite{galhotra2022causal}. Since the intervention fairness is too restrictive, this method identifies the fairness features by artificially specifying admissible features~\cite{salimi2019interventional}. However, while this method partially addresses the issue of the statistic-based fair feature selection algorithm's inability to explain the causal relationship between features, it poses challenges due to the lack of clear criteria for specifying admissible features. Different choices of admissible features can result in nominally fair features that may be considered unfair, and the feature selection results are unreliable, as shown in Figure 1. This challenge arises because the admissible feature-based method can only elucidate the causal relationship between a feature and the sensitive variable under specific conditions, leaving the causal relationship under different conditions unexplained. Moreover, the fair features identified by this method may not necessarily be relevant to the class variable, resulting in decreased accuracy of the trained classification model.
In conclusion, this method can not fully elucidate the causal relationship between features and sensitive variables by artificially specifying admissible features. Consequently, there remains a possibility of sensitive information being transmitted to the decision-making model, rendering the results unreliable.

Unlike traditional feature selection, causal feature selection explains the complete causal relationship among features, including the class variable~\cite{yu2020causality}. Causal feature selection aims to identify the Markov Blanket (MB) of a class variable, and the MB of a variable includes its parents (direct causes), children (direct impacts), and spouses (other direct causes of direct effects) in a faithful Bayesian network (BN)~\cite{aliferis2010local}. Given the MB of a target variable, all other variables become independent of that specific variable, and the MB of a class variable is the optimal subset of features for classification prediction tasks and offers good interpretability. However, the existing causal feature selection methods do not consider the impact of sensitive information on fairness.

Considering that the existing fair feature selection algorithms cannot fully explain the causal relationship between features and sensitive variables, and cannot effectively prevent the transmission of sensitive information, the main contributions of this paper are as follows:

\begin{itemize}

\item We introduce a fair causal feature selection problem that adheres to causal fairness. Additionally, we theoretically analyze the challenges of obtaining the fair causal feature set in this problem.

\item We propose a fair causal feature selection algorithm called FairCFS, which uses the MB of sensitive variables to block the path from features to sensitive variables, to identify a feature subset within the class variable's MB for ensuring fairness.

\item We conduct experiments on seven real-world datasets to show that FairCFS has comparable accuracy but achieves higher fairness than six causal feature selection algorithms and two fair feature selection algorithms.
\end{itemize}

The remaining sections of this paper are structured as follows: Section 2 reviews relevant work. Section 3 presents the fundamental definitions. In Section 4, we introduce the concept of fair causal feature selection and offer a theoretical analysis of the problem's maximum solution. Section 5 introduces the FairCFS algorithm, provides algorithmic correctness proofs, and offers detailed analysis. Section 6 presents the experimental results and corresponding analysis. Finally, we summarize the paper and outline potential avenues for future research in Section 7.

\section{Related work}

This paper aims to enhance the fairness of feature selection by incorporating considerations of causal relationships between features and constructing machine learning models that are more interpretable and socially acceptable. Thus, this section introduces the work related to the causal feature selection algorithm in section 2.1 and the related work of the fairness algorithm in section 2.2.

\subsection{Causal feature selection}

An emerging method for causal feature selection involves the identification of the MB associated with class attributes as a subset of features~\cite{aliferis2010local}. As illustrated in the introduction, class variables exhibit statistical independence from all other features when conditioned on MB. Consequently, MB explicitly identifies class attributes and establishes a local causal relationship between class attributes and features, a feature absent in traditional feature selection methods. Currently, causal feature selection methods are predominantly categorized as synchronous and divide-and-conquer methods~\cite{aliferis2010local,guyon2007causal}.

The synchronous algorithm represents the initial approach in the development of causal feature selection algorithms. The Growth-Shrink (GS)~\cite{margaritis1999bayesian} algorithm is the first correct synchronous algorithm. In its operation, it explores all candidate MB variables during the growth phase and subsequently prunes false-positive MB variables during the shrinkage phase. Subsequent advancements by Tsamardinos and Aliferis led to the enhancement of the GS algorithm through the introduction of incremental association MB (IAMB)~\cite{tsamardinos2003algorithms}. Following the introduction of IAMB, several variants emerged, including inter-IAMB~\cite{tsamardinos2003algorithms}, fast-IAMB~\cite{yaramakala2005speculative}, and FBED~\cite{borboudakis2019forward}. While the synchronous algorithm has proven effective, it is essential to note that the number of required data samples grows exponentially with the MB size of the target variable.

To reduce the amount of data required, the researchers proposed a divide-and-conquer algorithm. Representative algorithms are MinMax MB (MMMB)~\cite{tsamardinos2003time}, HITONMB~\cite{aliferis2003hiton}, Parent and Child-Based MB algorithm (PCMB)~\cite{pena2007towards}, Parent and Child-based MB Iterative Search algorithm (IPCMB)~\cite{fu2008fast}, etc. These algorithms are based on the GLL framework, which first finds the parent and child node PC of the target variable and then looks for the spouse node of the target variable from the PC node of the PC node. However, the spouse obviously cannot be obtained from the PC of the variables in the combination of the target parents and some children (only one parent), so there is no need to discover the PC of these variables~\cite{9783043}. Therefore, a new causal feature selection framework, called CFS~\cite{9783043}, with effective mate discovery is proposed to improve the efficiency of causal feature selection. The practice is to identify children with multiple parent nodes in the PC of target variable, then perform PC discovery only on these children to reduce the number of variables required for PC discovery.

In general, while existing causal feature selection algorithms can effectively explain causal relationships between variables, they do not consider algorithmic fairness, and the selected MB feature sets may lead to discrimination.

\subsection{Fair machine learning}

Fair machine learning algorithms represent a dedicated research domain to address inequity and bias within machine learning models~\cite{pessach2022review}. The significance of fairness in upholding human rights, ethical standards, and social justice has gained increasing prominence. Ensuring fair decision-making for diverse groups is essential for bolstering the credibility of technology and fostering the sustainable development of society. Many recent papers have proposed ways to enhance fairness in machine learning algorithms, and these methods are generally divided into three types: pre-processing, in-processing, and post-processing~\cite{pessach2022review}.

Pre-processing mainly consists of changing the training data before entering the machine learning algorithm. Early preprocessing methods such as Kamiran and Calders ~\cite{kamiran2012data} and Luong ~\cite{luong2011k} suggest changing the labels or reweighting certain instances before training to make the classification fairer. In general, samples closer to the boundaries of the decision are prone to change labels because they are most likely to be distinguished. Recent methods suggest modifying the data feature representation so that subsequent classifiers are fairer~\cite{calmon2017optimized}.

The in-processing method mainly modifies the machine learning algorithm to consider fairness during training~\cite{zafar2017fairness}. For example, Zafar~\cite{zafar2017fairness} et al. and Woodworth~\cite{pmlr-v65-woodworth17a} et al. suggest adding constraints to a classification model that needs to satisfy an indicator of equilibrium odds or other influences~\cite{zafar2017fairness}. Bechavod and Ligett ~\cite{bechavod2017learning} recommend incorporating penalties into an objective function that enforces the satisfaction of the metrics for FPR and FNR. Zemel~\cite{pmlr-v28-zemel13} et al. combine fair representation learning with process models by applying a logistic regression-based multi-objective loss function, and Louizos ~\cite{louizos2015variational} et al. apply this idea using variational autoencoders.

The post-processing method mainly modifies the output score of the classifier to make the decision fairer ~\cite{corbett2017algorithmic}. For example, Hardt~\cite{hardt2016equality} et al. propose a technique for transforming certain decisions of classifiers to enhance equilibrium odds or chances. Corbett-Davies~\cite{corbett2017algorithmic} et al. and Menon and Williamson ~\cite{pmlr-v81-menon18a} recommend choosing separate thresholds for each group to maximize accuracy and minimize demographic parity. Dwork~\cite{dwork2017decoupled} et al. propose a decoupling technique to learn different classifiers for each group, and they also combine transfer learning techniques with programs that learn from out-of-group samples.

In summary, while existing fairness algorithms have proposed many fair practices, there is still a lack of appropriate approaches to explain the causal relationships between features. Based on the above work, in this paper, we attempt to combine the fairness algorithm with the causal feature selection algorithm, and propose a machine learning algorithm that is both interpretable and fair.

\section{Definitions}
In this section, we delve into algorithmic fairness and explore the definitions and theorems of causality. Before delving into the definition of causal fairness, it is essential to provide some context by introducing the causal diagram.

\subsection{Causal diagram}

In this subsection, we provide essential background knowledge related to causal discovery. This paper employs a causal graph model based on Bayesian networks to represent causal relationships between variables, a well-established approach for comprehending causality. Let $P(V)$ denote the joint probability distribution across the set of all variables $V$, and $G = (V, E)$ represents a directed acyclic graph (DAG) consisting of nodes $V$ and edges $E$, often referred to as a causal graph. Each edge signifies a direct dependency between two variables within the DAG. Within the DAG framework, the notation $V_{i}\rightarrow V_{j}$ indicates that $V_{i}$ serves as the parent of  $V_{j}$, and conversely, $V_{j}$ is the child of $V_{i}$, indicating that $V_{i}$ exerts an influence on $V_{j}$.

\begin{definition}
\textbf{(Bayesian Network)}~\cite{pearl1988morgan} A triplet $< V, G, P(V) >$ is called a Bayesian network iff $< V,G,P(V) >$ satisfies the Markov condition: Given the parents of a variable, each variable in $V$ is independent of all non-descendants of the variable.
\end{definition}
A BN encodes the joint probability over a set of variables $V$ and decomposes $P(V)$ into the product of the conditional probability distributions of the variables given their parents in $G$. Let $Pa(V_{i} )$ be the set of parents of $V_{i}$ in $G$. Then, $P(V)$ can be written as:

\begin{equation}\nonumber
    P(V_{1}, V_{2}, . . . , V_{n+1} ) = \prod_{i=1}^{n+1}P(V_{i} \mid Pa(V_{i})).
\end{equation}

\begin{definition}
\textbf{(Faithfulness)}~\cite{spirtes2000causation} A $BN < V, G, P >$ is faithful iff all conditional dependencies between features in $G$ are captured by $P$.
\end{definition}

Faithfulness indicates that in a BN, $X$, and $Y$ are independently conditioned on a set $S$ in $P$ iff they are d-separated by $S$ in $G$.

\begin{definition}
\textbf{(Markov Blanket)}~\cite{pearl1988morgan} In a faithful BN, each variable has only one MB consisting of its parents, children, and spouses (the other parents of its children).
\end{definition}

Given the MB of $Y$, $MB_{Y}$, all other variables are conditionally independent of $Y$, that is,
\begin{equation}\nonumber
 X \!\perp\!\!\!\perp Y\mid MB_{Y}, \forall  X \in V \backslash MB_{Y} \backslash \{X\}.
\end{equation}

Pearl introduced the concept of intervention, which involves altering the state of attributes to a specific value and observing the effects.
\begin{definition}
\textbf{(Do-operator)}~\cite{pearl2009causality}  An intervention on an attribute $X$, denoted as  $X\leftarrow x$, is effectively achieved by assigning the value $x$ to the variable $X$ within a modified causal graph $G'$, $G'$  is identical to $G$, except all incoming edges to $X$ have been eliminated.
\end{definition}

 The do-operator aligns with the graphical interpretation of an intervention. Specifically, an intervention represented as $do(X) = x$ equals conditioning on $X = x$ when $X$ lacks any ancestors within $G$.

\subsection{Causal fairness}
This paper aligns with interventional fairness, a concept within the realm of causal fairness. Causal fairness solves the challenges of statistical fairness, which primarily relies on correlation-based assessments. Predictive algorithms often struggle to differentiate between causal relationships and spurious correlations among attributes. A well-known illustration of this is Simpson's paradox~\cite{wagner1982simpson}:

\begin{example}
In the 1973 Berkeley admissions event, the acceptance rate for male applicants was 44\%, and the acceptance rate for female applicants was 35\%. While there may be a bias against women, when the data is broken down by department, women's acceptance rate is higher than that of men in each college. This is because most women apply to colleges with strict admission requirements, the overall female acceptance is underestimated.
\end{example}

This example underscores the importance of delving into causal relationships between sensitive attributes and decisions instead of relying solely on correlations to tackle equity concerns effectively~\cite{makhlouf2020survey}. In this paper, we adopt the Interventional Fairness concept formulated by Babak~\cite{salimi2019interventional} et al. This robust causal fairness concept can be assessed using input datasets and accurately captures group-level fairness. It ensures that the sensitive variables $S$ does not influence the output $O$ when other variables are held constant at arbitrary values.

\begin{definition}
\textbf{(K-fair)}~\cite{salimi2019interventional} Fix a set of attributes $K \subseteq V-\{S, O\}$. We say that an algorithm $\ell:$ Dom(X) $\rightarrow$ Dom(O) is K-fair w.r.t. a sensitive attribute $S$ if, for any context $K = k$ and every outcome $O = o$, the following holds:

$Pr(O = o \mid do(S = 0), do(K = k)) = Pr(O = o \mid do(S = 1), do(K = k)).$
\end{definition}
If the algorithm is K-fair for each set $K$, the algorithm is said to be intervening fairly.
Moreover, in the intervened figure $G'$ (the incoming edge from $S$ to $K$ is removed), the sensitive attribute $S$ is independent of $Y'$ under $K$ conditions, i.e., $S$ and $Y'$ in figure $G'$ are d separated under $K$ conditions.

\section{Problem statement}

With the definition of K-fair, this section establishes a fair causal feature selection problem that guarantees intervention fairness. Fair causal feature selection aims to find a feature subset that, while satisfying the concept of fairness, also possesses causal relationships that can be explained. We provide a theoretical analysis to explain this problem.

\subsection{Problem analysis}

Consider dataset $D$, $V=S \cup X \cup Y $, sensitive variables $S$, non-sensitive variables $X = {X_{1}, X_{2},..., X_{n}}$, label variable $Y$. $MB_{Y}$, $MB_{S}$ are the collection of Markov blanket variables for $Y$ and $S$, containing parent-child and spouse nodes for $Y$ and $S$, respectively. $Y'$ is the target variable obtained after training on the subset $T \subseteq V$. Now, we propose a definition of fair causal features:

\begin{definition}
\textbf{(Fair Causal Features)} Consider a set of features $T$ as fair causal features if (1) the predictor of the classifier model $Y'$ trained on $T$ satisfies K-fair, and (2) the features in $T$ are interpretable with the class variable $Y$.
\end{definition}
Here $T \subseteq V$. The objective of feature selection is to identify the fair subset $T$ after recognizing the causal relationships between features and both the class variable and sensitive variable, and this subset ensures that training $Y'$ using these variables results in causal fairness.

\begin{example}
For example, a loan system that decides whether to approve an applicant's loan application based on their personal information. In this example, we can see that credit score, annual income, education level, and marital status are potential fair features, while gender and age can affect the fairness of loan approval decisions. Following the concept outlined in Definition 6, when selecting fair causal features, gender is initially excluded as it is considered a sensitive attribute. Other features such as credit score and annual income may be chosen as fair causal features because they can predict an individual's repayment capability without involving gender discrimination. Conversely, age, as a feature that doesn't adhere to causal fairness, is also excluded from consideration.
\end{example}

This paper aims to find all fair causal features that can be used for classifier training without specifying admissible features, with the sensitive feature $S$ excluded from training. Therefore, we decide to search for the maximum target subset in the set of the class variable's MB and determine the conditional independence between features and the sensitive variable based on the MB of the sensitive variable. According to Definition 3, first discovering the MB of the class variable ensures that the selected features have predictive information for the target variable $Y$, improving prediction accuracy. Secondly, judging the conditional independence between features and the sensitive variable under the condition of the MB of the sensitive variable can be provided by the algorithm to identify the set of features that block the paths from fair features to the sensitive variable without the need to specify admissible features manually. Below, we will define the fair causal feature selection problem:

\begin{definition}
\textbf{(Fair Causal Feature Selection Problem)} Given a dataset $D=\{X,S,Y\}$, identify the largest subset $T \subseteq MB_{Y}$, where $T$ has an explainable causal relationship with $Y$ while ensuring that the feature set $T$ is causally fair, where $MB_{Y} \subseteq X$.
\end{definition}

Based on the definition of causal fairness (Definition 5), we need to obtain the distribution after intervening on the causal graph when we want to determine if a feature set meets the causal fairness criteria defined in Definition 6. However, in practical training, this can often be challenging to obtain. Building upon the assumption proposed by Galhotra~\cite{galhotra2022causal}, we will introduce a method to assess the fairness criteria defined in Definition 6 using the assumed intervention distribution through classifier training. This approach separates the fairness of feature selection from the training process. It allows us to theoretically analyze how to eliminate bias effects during the feature selection process without considering actual intervention operations.

\textbf{Classifier Training: }A new variable $Y'$ (predictive variable) is generated by training a classifier on the selected feature subset $T$. Here, $Pr(Y' \mid T)$ is derived from the observed distribution $P(V)$, which is equivalent to adding a new node $Y'$ to the causal graph. $Y'$ is a child node of all features that can affect the classifier's output.

\begin{assumption}
~\cite{galhotra2022causal} The mechanism generating $Y'$ is the same as $P(Y' \mid A \cup T)$, where $P(\cdot)$ is the observational distribution.
\end{assumption}

In Fair Causal Feature Selection Problem, $A=\emptyset$. The problem satisfies the assumption.

According to the definition of causal fairness, whenever we intervene with the fair feature set, the predictive algorithm's output distribution remains unchanged when sensitive variable values change. Using the do-operation, intervening with a feature is equivalent to removing its incoming edges and adjusting its value. Suppose all paths from the sensitive variable to the predictive target $Y'$ through the algorithm-selected features are blocked after intervention under the sensitive variable's MB condition set. In that case, the features chosen by the algorithm are considered fair.

Therefore, finding a solution to the Fair Causal Feature Selection Problem requires adequately explaining the causal relationships between features and label. Since all paths from fair features to sensitive variables are blocked under the condition set $MB_{S}$, the causal relationships of every path from fair features to sensitive variables are identified. Furthermore, since $T\subseteq MB_{Y}$, fair features are members of the MB of the class variable, establishing a dependency relationship with the class variable. This means that the causal relationships of every path, from fair features to the class variable, are also identified. Based on Definition 2, the variable's MB is unique, and the transmission of sensitive information is blocked. There will not be a situation where fair causal features are identified as unfair when the condition set changes.

We point out through the following Lemma 1 that the Fair Causal Feature Selection Problem can find the maximum solution with the most features by merging the sets that satisfy Fair Causal Features.

\begin{lemma}
If two different sets of Fair Causal Features, $D_{1}$, and $D_{2}$, are solutions to the Fair Causal Feature Selection Problem, then the new set after merging $D_{1}$ and $D_{2}$ is still a causally fair solution. The classifier trained on $D_{1} \cup D_{2}$ is also causally fair.
\renewcommand\qedsymbol{\ensuremath{\blacksquare}}

\begin{proof}
Let $Y_{1}'$ and $Y_{2}'$ denote the output variables of the classifier trained on $D_{1}$ and $D_{2}$. Let $G'$ denote a modified causal graph where the incoming edges of $S$ are removed. According to the definition of causal fairness, all paths from the sensitive feature $S$ to $Y_{1}'$ are blocked in $G'$, i.e., $ S \!\perp\!\!\!\perp Y_{1}' \mid G'$. Since $Y_{1}'$ is a child of attributes in $D_{1}$, all paths from $S$ to the parents of $Y_{1}'$ are blocked, i.e., $ S \!\perp\!\!\!\perp pa(Y_{1}') \mid G'$. We obtain the same condition for $D_{2}$. Let $Y'$ denote the output variable of the classifier trained on $D_{1} \cup D_{2}$. We have:
\begin{equation}\nonumber
\begin{aligned}
&Pr [Y' \mid do(S)=s, MB_{S} ]\\
&= \Sigma_{pa(Y')=c}   Pr[Y'=y \mid pa(Y')=c, do(S)=s, MB_{S} ] Pr[pa(Y')=c \mid do(S)=s, MB_{S}]\\
&= \Sigma_{pa(Y')=c}   Pr[Y'=y \mid pa(Y')=c, MB_{S}] Pr[pa(Y')=c \mid do(S)=s, MB_{S}]\\
(i)\\
&= \Sigma_{pa(Y')=c}   Pr[Y'=y \mid pa(Y')=c, MB_{S}] Pr_{G'} [pa(Y')=c \mid S=s, MB_{S}]\\
(ii)\\
&= \Sigma_{pa(Y')=c}   Pr[Y'=y \mid pa(Y')=c,MB_{S}] Pr_{G'} [pa(Y')=c \mid MB_{S}]\\
\end{aligned}
\end{equation}

(i) Because performing a do-operation on $S$ is equivalent to creating a new causal graph $G'$, where the value of $S$ is set to $s$, $Y'$ only has parent nodes $D_{1} \cup D_{2}$, hence $Pr[pa(Y')=c \mid do(S)=s, MB_{S}] = Pr_{G'} [pa(Y')=c \mid S=s, MB_{S} ]$.

(ii) Since $Y'$ is trained over $D_{1} \cup D_{2}$, $pa(Y') \subseteq D_{1} \cup D_{2}$, $ S \!\perp\!\!\!\perp pa(Y') \mid G'$, $Pr_{G'}$[$pa(Y')=c \mid$ S=s, $MB_{S}$]=$Pr_{G'}$[$pa(Y')=c \mid MB_{S}$].

\end{proof}
\end{lemma}
In $G'$, $Pr_{G'} [pa(Y')=c \mid S=s, MB_{S}] = Pr_{G'} [pa(Y')=c \mid MB_{S}]$, $Y'$ satisfies Definition 5, therefore, $D_{1} \cup D_{2}$ is causally fair.

According to the following Lemma 2, the Fair Causal Feature Selection Problem has a maximum solution. This means that we need to find a unique maximal set that satisfies the requirements of causal fairness when selecting a feature set. The proof for this is as follows:

\begin{lemma}
Fair Causal Feature Selection Problem has the maximum solution $D$.
\renewcommand\qedsymbol{\ensuremath{\blacksquare}}

\begin{proof}
Suppose the problem does not have a maximum solution. Let $D_{1} \cup D_{2}$ be two different maximal sets of features that ensure causal fairness. Using Lemma 2, $D_{1} \cup D_{2}$ also ensures causal fairness. Since $D_{1}$ $\neq$ $D_{2}$, $|D_{1} \cup D_{2}| > |D_{1} |$, $|D_{2} |$. This is a contradiction, as $D_{1} \cup D_{2}$ are maximal sets. Therefore, the assumption that the problem does not have a maximum solution is wrong.
\end{proof}
\end{lemma}

 We conclude that the Fair Causal Feature Selection Problem has a maximum solution, ensuring that the fair causal feature set satisfies causal fairness and possesses predictive information for $Y$.
%and has an explainable causal relationship.

\section{Fair Causal Feature Selection Algorithm}

In this Section, we first provide the FairCFS algorithm and gradually prove the theoretical correctness of FairCFS in Section 5.1, and then we conduct an analysis of FairCFS through an example in Section 5.2.

\subsection{Algorithm Implementation}
The FairCFS algorithm utilizes existing MB discovery algorithms to find the MBs of the class and sensitive variables (the sensitive variable is known). Then, according to Lemma 1 and Lemma 2, through conditional independence tests to assess the independence between features, it selects suitable features that meet the requirements for causal fairness and have an explainable causal relationship with $Y$. Below, we provide the FairCFS algorithm step by step.

\begin{algorithm}[t]
\caption{FairCFS algorithm}
\label{alg:algorithm}
\setstretch{1.15}

\begin{flushleft}
\ \ \textbf{Input}: $S$: Sensitive feature; $Y$: Label\\
\ \ \textbf{Output}: $[M_{1} \cup M_{2} ]$: Fair causal features
\end{flushleft}

\begin{algorithmic}[1]
\STATE /*Step 1: Discover Markov blankets of $Y$ and $S$*/
\STATE $MB_{Y} \leftarrow GetMB(Y);$
\STATE $MB_{S} \leftarrow GetMB(S);$
%\STATE $MB_{Y}, MB_{S} \leftarrow GetMB(Y), GetMB(S);$
\STATE /*Step 2: Searching for features independent of $S$ from $\{MB_{Y} \setminus MB_{S}\}$*/
\STATE \textbf{for} each\ $X\in \{MB_{Y} \setminus MB_{S}\}$\ \textbf{do}
\STATE \quad\textbf{if} ${X \!\perp\!\!\!\perp S\mid MB_{S}}$ \ \textbf{then}
\STATE \quad\quad$M_{1} \leftarrow M_{1} \cup X;$
\STATE /*Step3: Searching for features independent of $S$ from $\{MB_{Y} \cap MB_{S}\}$ */
\STATE \textbf{for} { each\ $X\in \{MB_{Y} \cap MB_{S}\}$}\ \textbf{do}
\STATE \quad\textbf{if} ${X \!\perp\!\!\!\perp S\mid Z}$ for some $Z\subseteq \{MB_{S}\setminus X\}$\ \textbf{then}
\STATE \quad\quad$M_{2} \leftarrow M_{2} \cup X;$

%\STATE $FairMB \leftarrow M_{1} \cup M_{2}; $
\STATE \textbf{return} $[M_{1} \cup M_{2}];$
%\STATE \textbf{return} $[M_{1} \cup M_{2}];$
\end{algorithmic}
\end{algorithm}

Step 1: In this step, we use the existing MB discovery algorithm, GetMB, to find the MB of the class variable $Y$ and the MB of the sensitive variable $S$. According to Definition 3, all other variables are conditionally independent of the $Y$ variable given $MB_{Y}$. Similarly for $S$. To initially ensure the model's accuracy after feature selection, in the subsequent steps, we only evaluate whether the variables in the MB set of the class variable $Y$ satisfy the definition of causal fairness features. This is because, in theory, the MB of the class variable $Y$ has been proven to be the optimal set for predicting classifications.

Step 2: For the features identified in Step 1 as part of $\{MB_{Y} \setminus MB_{S}\}$, we individually assess their conditional independence with the sensitive variable $S$. If a feature, when conditioned on $MB_{S}$, is independent of $S$, according to the following Lemma 3, any feature with this property will block the transmission of sensitive information along all paths to the sensitive variable. This ensures causal fairness and satisfies our definition of fair causal features. The proof is as follows:

\begin{lemma}
Consider a dataset $D$ with sensitive $S$ and a collection of features $M_{1}$ and the MB of $S$, if $M_{1} \!\perp\!\!\!\perp S\mid MB_{S}$, then $M_{1}$ is causally fair.
\renewcommand\qedsymbol{\ensuremath{\blacksquare}}

\begin{proof}
Given $M_{1} \!\perp\!\!\!\perp S\mid MB_{S}$ for $MB_{S}$, the feature $M$ does not capture any information about the sensitive variables. Hence, all paths from $S$ to the target $Y'$ that pass through $M_{1}$ are blocked. Mathematically, We have:
\begin{equation}\nonumber
\begin{aligned}
&Pr [Y' \mid do(S),
 MB_{S} ]= \Sigma_{M_{1}}  Pr[Y' \mid M_{1}, do(S), MB_{S} ]Pr[M_{1} \mid do(S), MB_{S}]\\
(i)\\
&= \Sigma_{M_{1}}  Pr[Y' \mid M_{1}, do(S), MB_{S}] Pr[M_{1} \mid MB_{S}]\\
(ii)\\
&=Pr[Y' \mid MB_{S}]\\
\end{aligned}
\end{equation}

(i) Since $M_{1} \!\perp\!\!\!\perp S\mid MB_{S}$, all paths from $M_{1}$ to $S$ are blocked. A classifier trained using $M_{1}$ will not capture any sensitive information about $S$, as sensitive information will not pass through $M_{1}$. Additionally, performing a do-operation on $S$ is equivalent to removing the incoming edges from $S$ to other nodes in the causal graph.

(ii) Based on the assumption about the construction of $Y'$ (Assumption 1), the variable $Y'$ is only dependent on the variables in $M_{1}$ in all environments. Given $M_{1}$, the variable $Y'$ is independent of $S$. Also, $S$ nodes do not have any incoming edges. Therefore, on applying the rule of do-calculus, since $Y'$ is independent of $S$ in the modified graph where, incoming edges of $S$ nodes that are ancestors of $M_{1}$ are removed. Thus, $Pr[Y' \mid M_{1}, do(S), MB_{S}] = Pr[Y' \mid M_{1}, MB_{S}]$.
%(ii) Using Lemma 1.
\end{proof}
\end{lemma}
This indicates that any intervention on $S$ will not affect the variable $Y'$, thus $Y'$ satisfies Definition 5. This ensures that the selected features $M_{1}$ are fair causal features.

Combining Lemma 3 with example 2 in the loan system tells us that if a feature set $M$ is conditionally independent of $S$ given $MB_{S}$, then these features are causally fair. In our loan system example, this means that if a feature, such as annual income, is independent of gender under the condition of considering the gender's MB, then this yearly income feature will not be used to discriminate or show bias against gender because its values are not influenced by gender.

Step 3: Even if a feature $M_{2}$ is found in the second step 2 dependent on the sensitive variable S, it may still be conditionally independent of S under the condition set Z, $Z \subseteq MB_{S}$. For example, if $M_{2}$ is the spouse node of the sensitive variable $S$, under the intervention of the sensitive variable $S$, the paths from the sensitive variable through this feature to $Y$ are also blocked, preventing the transmission of sensitive information. Therefore, this feature also complies with the definition of fair causal features and can be added to the selected feature set. At this point, we have found all feature sets $M_{1} \cup M_{2}$ that satisfy the causal fairness criteria.

\begin{lemma}
Consider dataset $D$ with sensitive $S$ and the $MB_{S}$ of $S$ and $Y$, a collection of features $M_{1}$  satisfying $M_{1} \!\perp\!\!\!\perp S\mid MB_{S}$ and a collection of features $M_{2}$  satisfying $M_{2} \not\!\perp\!\!\!\perp S\mid MB_{S}$, but $M_{2} \!\perp\!\!\!\perp S\mid Z$, for some $Z \subseteq MB_{S}$, $M_{2}$ is causally fair.
\renewcommand\qedsymbol{\ensuremath{\blacksquare}}

\begin{proof}
Given $M_{2} \not\!\perp\!\!\!\perp S\mid MB_{S}$, but $M_{2} \!\perp\!\!\!\perp S\mid Z$, for some $Z \subseteq MB_{S}$, according to Lemma 4, the feature $M_{2}$ does not capture any information about the sensitive variables under the condition $Z$. Hence, all paths from $S$ to the target $Y'$ that pass through $M_{2}$ are blocked. We have:
\begin{equation}\nonumber
\begin{aligned}
&Pr [Y' \mid do(S), Z] = \Sigma_{M_{2}} Pr[Y' \mid X, do(S), Z]Pr[X \mid do(S), Z]\\
(i)\\
&= \Sigma_{M_{2}} Pr[Y' \mid X, Z] Pr[X \mid do(S), Z]\\
(ii)\\
&=Pr[Y' \mid Z]\\
\end{aligned}
\end{equation}

(i) Since all paths from $M_{2}$ to $S$ are blocked, a classifier trained using $M_{2}$ will not capture any sensitive information about $S$, and $M_{2} \!\perp\!\!\!\perp S\mid Z$ holds.

(ii) Based on the assumption about the construction of $Y'$ (Assumption 1), the variable $Y'$ is only dependent on the features in $M_{2}$ in all environments. Given $M_{2}$, the variable $Y'$ is independent of $S$. Also, $S$ nodes do not have any incoming edges. Therefore, on applying the rule of do-calculus, since $Y'$ is independent of $S$ in the modified graph where, incoming edges of $S$ nodes that are ancestors of $M_{1}$ are removed. Thus, $Pr[Y' \mid M_{2}, do(S), Z] = Pr[Y' \mid M_{2}, Z]$.
%The prediction result $Y'$ obtained through $M_{2}$, $M_{2} \!\perp\!\!\!\perp S\mid Z$ have $Pr[Y' \mid Z]=Pr[Y']$.

Consequently, $M_{2}$ is the set of fair causal features.
\end{proof}
\end{lemma}
%Lemma 5 introduces a very critical case where specific feature sets $M_{2}$ can still be considered causally fair when they are independent of the sensitive variable S under the selected $M_{1}$ condition but dependent on the sensitive attribute S under the $MB_{S}$ condition. For example, marital status may belong to $M_{2}$ in a loan approval system. It may be independent of the sensitive attribute S under the selected $M_{1}$, but there might be some association between marital status and gender. However, when the fairness feature, such as income, is known in specific social contexts, the potential for marital status to introduce discrimination in loan decisions might disappear. Therefore, according to Lemma 5, as long as the paths through $M_{1}$ are blocked when intervening on gender, marital status can still be considered a causally fair feature.

Through Theorem 1, we summarize the feature objectives chosen by the FairCFS algorithm to address the problem of fair causal feature selection as follows:

\begin{theorem}
\renewcommand\qedsymbol{\ensuremath{\blacksquare}}
Consider dataset $D$ with sensitive $S$, a set of features $X$ with a target $Y$. A set $X\in MB_{Y}$ is safe to be added along with $T$, where $T \subseteq M$  without violating causal fairness iff $M \!\perp\!\!\!\perp S\mid Z$ for some $Z \subseteq MB_{S}$.

\begin{proof}
Using Lemma 1, 3, and 4, we can observe that all the features $X \in M_{1} \cup M_{2}$ such that $M_{1} \!\perp\!\!\!\perp S\mid MB_{S}$, and $M_{2} \not\!\perp\!\!\!\perp S\mid MB_{S}$, but $M_{2} \!\perp\!\!\!\perp S\mid Z$, for some $Z \subseteq MB_{S}$, are safe to be added without worsening the fairness of the dataset.
\end{proof}
\end{theorem}

\subsection{Algorithm analysis}
In this section, we will introduce the specific objectives of the FairCFS algorithm for feature selection based on a detailed example graph. Unlike existing fairness feature selection methods that rely on user-selected admissible variables to ensure fairness but suffer from the problem of unclear criteria for specifying admissible features, which leads to unreliable results and potentially irrelevant features, FairCFS takes a different approach. It does not specify admissible variables but instead focuses on finding the condition set that blocks the transmission of sensitive information by discovering the Markov blankets of both the sensitive and class variables. Additionally, $MB_{Y}$ is the optimal feature subset for class variable classification, and FairCFS ensures the accuracy of its classification model.

\begin{figure}[t]
  \centering
  \includegraphics[height=2.0in]{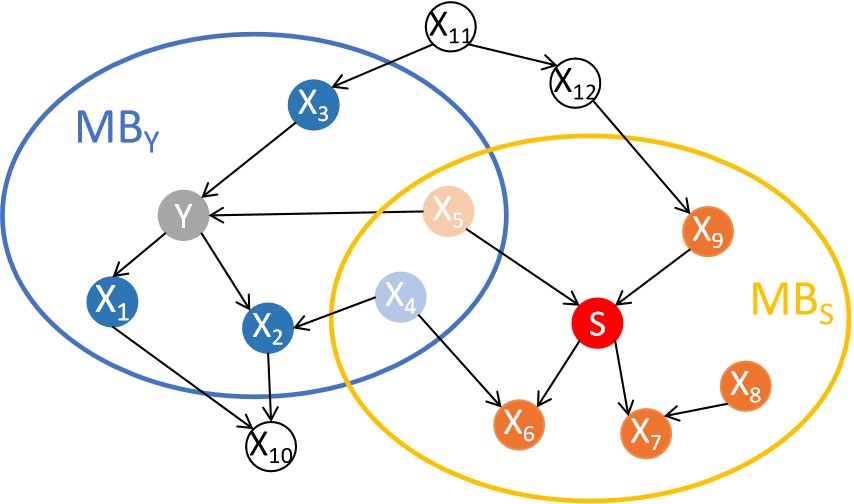}\\
   \caption{\textbf{ A local causal diagram constructed by FairCFS to select fair causal features.} (1) The features $X_{1} - X_{4}$ all have a condition set that blocks all their paths to $S$, that is, these features are independent of $S$ in the case of condition $Z$, where $Z \in MB_{S}$. These features are then selected as fair causal features; (2) Features $X_{6} - X_{9}$ could not find the set of conditions that blocked it to the sensitive variable S, that is, they are not fair causal features; (3) Although the $X_{10}- X_{12}$ also satisfy independent of $S$ in the case of some condition $Z$, they are not features in $MB_{Y}$, and will not be selected. }
\end{figure}

Figure 3 illustrates a local causal graph of the scenario example, including the class variable $Y$, the sensitive variable $S$, the variables in their respective MB, and others. Below, we will analyze the objectives of FairCFS in selecting features and the situations where certain features are not chosen, using this example and the algorithm process.

\begin{itemize}

\item After constructing the local causal graph, we can see that features $X_{1}, X_{2}$, and $X_{3}$ in the diagram correspond to the variables $M_{1}$ in algorithm step 2. All paths from these variables to $S$ are blocked by $MB_{S}$, preventing sensitive information from being transmitted. According to Lemma 3, they do not contain information related to the sensitive variable and can be used for training to ensure fairness. Using the $G^{2}$ conditional independence test, we can confirm that these features are independent of $S$ under the condition of $MB_{S}$, i.e., $M_{1} \!\perp\!\!\!\perp S\mid MB_{S}$.

\item After selecting the features in $M_{1}$, it is observed that $X_{4} \not\!\perp\!\!\!\perp S\mid MB_{S}$, thus $X_{4}\not\in M$. However, the path from $X_{4}$to $S$ can be blocked under a subset $Z$ of the existing $MB_{S}$, meaning that $X_{4} \!\perp\!\!\!\perp S\mid Z$, for some $Z \subseteq MB_{S}$. Furthermore, we do not use the sensitive variable $S$ (equivalent to intervening on $S$ and removing its incoming edge to $Y$), sensitive information will not be transmitted through $X_{4}$. According to Lemma 4, this feature can be used, and $X_{4}$ also corresponds to the feature $M_{2}$ in the algorithm, i.e., $M_{2} \!\perp\!\!\!\perp S \mid Z$ for some $Z \subseteq MB_{S}$.

\item The analysis of features that are not selected is as follows: features $X_{5}, X_{6}$, and $X_{7}$ in the graph also have all paths to $S$ blocked by $MB_{S}$, and using these features for training can ensure fairness. However, since these features are outside the $MB_{Y}$, it indicates that they do not contain information related to $Y$. Using these features to improve the accuracy of the classifier will not provide good results. Through the $G^{2}$ conditional independence test, it can be obtained that, under the condition of $MB_{Y}$, these features are independent of $Y$, These features are independent of $Y$ under the $MB_{Y}$ condition, then they are not causally fair.

\item Another type of feature that should not be selected like $S_{5}$, where $S_{5}\in MB_{Y}$, but there is no path from $S_{5}$ to $S$ that $MB_{S}$ can block. This indicates that using such variables would include discriminatory information and such features should not be selected. In other words, $S_{5} \not\!\perp\!\!\!\perp S $, $S_{5}$ is not causally fair.
\end{itemize}

The above points explain the distinguishing features of the FairCFS algorithm when selecting fair causal features. Ultimately, the selected features can all find a conditional set $Z$, which blocks the paths from these features to the sensitive variable $S$, preventing the transmission of sensitive information. Thus, using these features for classifier training will not lead to discrimination.

\section{EXPERIMENTS}
In this section, we used seven fair classification datasets to evaluate the algorithms FairCFS against six causal feature selection algorithms and two fair feature selection algorithms, respectively.

\subsection{Expermental Setup}
To validate the accuracy and fairness of the FairCFS algorithm, we conducted experiments on seven real-world datasets and compared them with six causal feature selection algorithms and two fair feature selection algorithms, respectively. The eight algorithms used for comparison are MMMB~\cite{tsamardinos2003time}, HITONMB~\cite{aliferis2003hiton}, PCMB~\cite{pena2007towards}, BAMB~\cite{ling2019bamb}, CFS~\cite{9783043}, STMB~\cite{gao2017efficient}, Auto~\cite{belitz2021automating}, and Seqsel~\cite{galhotra2022causal}. The first six algorithms are implemented in the existing MATLAB package\footnote{\url{http://bigdata.ahu.edu.cn/causal-learner}} Causal Learner~\cite{ling2022causal}. Additionally, we implemented FairCFS in MATLAB and compared it against these eight feature selection algorithms. All experiments were conducted on a computer running Windows 10 with an Intel Core i5-13490F CPU and 32GB of RAM. The significance level for the $G^{2}$ conditional independence test was set to 0.01~\cite{neapolitan2004learning}. The algorithm is as follows:

\begin{itemize}

\item \textbf{MMMB}: The MMMB algorithm is a causal feature selection algorithm, which first uses the MMPC algorithm to obtain the PC of the class variable, then learns the PC of the variable in the PC to get the spouses of the class variable.

\item \textbf{HITONMB}: In contrast to MMMB, HITONMB uses the HITONPC algorithm to learn the PC of variables.

\item \textbf{PCMB}: In contrast to MMMB, PCMB uses the GetPCD algorithm to learn the PC of variables.

\item \textbf{BAMB}: The BAMB algorithm alternately adds and remove the target variable for PC and spouses.

\item \textbf{CFS}: The CFS algorithm identifies children with multiple parent nodes in the PC of target variable, then identifies these children's PC to learn the spouses of target.

\item \textbf{STMB}: The STMB algorithm first uses the PCsimple algorithm to learn the PC of target variable, and then uses the coexisting property to learn the spouses of target.

\item \textbf{Auto}: The Auto algorithm first trains a classifier for each feature, then selects the features with the best AUC indicator to combine with the rest of the features, and retrains the classifier in the next round until the end of the 100-round cycle.

\item \textbf{Seqsel}: Based on the Rcit conditional independence test, Seqsel finds fair features based on whether features and class variables are independent under the condition of an admissible feature set.

\end{itemize}

\textbf{Datasets: }We used seven publicly available and commonly used datasets for fairness classification tasks to compare the results of the FairCFS algorithm with existing causal feature algorithms and fair feature selection algorithms. The details of these datasets are shown in Table 1. Based on a fair dataset survey~\cite{le2022survey}, we followed the procedures for handling attribute values, missing values, and sensitive feature selection.

\begin{table}[]
\setlength\tabcolsep{7pt}
\small
\centering
\centering\caption{Fair classification datasets}
\label{tab:my-table}
\begin{tabular}{llll}
\toprule
datasets               & Num.samples & Num.features & Sensitive feature \\
\midrule
Law                    & 20798    & 11           & race               \\
Oulad                  & 21562    & 10           & gender            \\
German                 & 1000     & 20           & age               \\
Compas                 & 6172     & 8            & gender            \\
CreditCardClients      & 30000    & 23           & gender            \\
StudentPerformanceMath & 395      & 32           & gender            \\
StudentPerformancePort & 649      & 32           & gender               \\
\bottomrule
\end{tabular}
\end{table}

\textbf{Classifiers and evaluation metrics}:
We applied the FairCFS and comparative algorithms to the datasets mentioned above to obtain the features selected by each algorithm. We uniformly trained classifiers, such as Logistic Regression (LR), Naive Bayes (NB), and k-nearest neighbors (KNN), on each dataset. We used ten-fold cross-validation for each dataset and evaluated the algorithms' accuracy and fairness using the following metrics:

\begin{itemize}

\item \textbf{Accuracy (ACC)}: Prediction accuracy is the percentage of correctly classified test samples in all samples. For ACC, larger values indicate that the model is more accurate.

\item \textbf{Statistical Parity Difference (SPD)}~\cite{dwork2012fairness}: SPD is used to measure the degree of difference in the classification results of a model between different groups (usually based on sensitive attributes such as gender or race). This measure is mathematically formulated as follows: $SPD = \mid P(Y' = 1 \mid S = s_{1}) - P(Y' = 1 \mid S = s_{2}) \mid$, $SPD \in [0,1]$. For SPD, smaller values indicate a fairer model.

\item \textbf{Predictive Equality (PE)}~\cite{corbett2017algorithmic}: This requires FPRs (meaning the probability of an individual with a negative outcome to have a positive prediction) to be similar across groups. This measure is mathematically formulated as follows:$PE = \mid P(Y' = 1 \mid S = 1, Y = 0) - P(Y' = 1 \mid S \neq 1, Y = 0) \mid$, $PE \in [0,1]$. For PE, smaller values indicate a fairer model.

\end{itemize}

\subsection{Comparison of FairCFS with causal feature selection algorithm}
\begin{table}[t]
\setlength\tabcolsep{5pt}
\small
\centering
\centering\caption{Comparison of FairCFS, CFS, BAMB, HITONMB, MMMB, PCMB, and STMB on LR Classifier ($\uparrow$ indicates that a higher value of the metric is better, while $\downarrow$ indicates that a lower value of the metric is better).}
\label{tab:my-table}
\begin{tabular}{cc||cccccccc}
\hline
metric &Algorithm    & German          & Compas          & Credit          & Law             & Oulad           & Studentm  & Studentp        \\
\hline
\multirow{7}{*}{ACC $\uparrow$} &CFS    & 0.7270 & 0.6762 & 0.8060 & \textbf{0.8921} & 0.6859 & 0.9188   & \textbf{0.9276}   \\
&BAMB    & 0.7280 & 0.6772 & 0.8060 & 0.8918 & 0.6859 & 0.9188   & 0.9259   \\
&HITONMB & 0.7280 & \textbf{0.6775} & \textbf{0.8070} & 0.8917 & 0.6860 & 0.9188   & 0.9275   \\
&MMMB    & \textbf{0.7290} & 0.6762 & 0.8069 & 0.8917 & \textbf{0.6869} & 0.9188   & 0.9244   \\
&PCMB    & 0.7280 & 0.6767 & \textbf{0.8070} & 0.8917 & \textbf{0.6869} & 0.9188   & 0.9244   \\
&STMB    & 0.7240 & \textbf{0.6775} & \textbf{0.8070} & 0.8918 & 0.6868 & 0.9112   & 0.9213   \\
&FairCFS     & 0.6980 & 0.5722 & 0.7788 & 0.8897 & 0.6796 & \textbf{0.9189}   & 0.9275    \\
\hline
\multirow{7}{*}{SPD $\downarrow$}        &CFS    & 0.1078          & 0.2409          & 0.0257          & 0.0058          & 0.0132          & \textbf{0.1298}          & 0.0597 \\
        &BAMB    & 0.1407          & 0.2226          & 0.0263          & 0.0055          & 0.0107          & 0.1298          & 0.0829          \\
&HITONMB & 0.2573          & 0.2321          & 0.0260          & 0.0093          & 0.0083          & 0.1298          & 0.0854          \\
&MMMB    & 0.2561          & 0.2368          & 0.0293          & 0.0091          & 0.0063          & 0.1298          & 0.0804          \\
&PCMB    & 0.1655          & 0.2375          & 0.0292          & 0.0091          & 0.0063          & 0.1298          & 0.0804          \\
&STMB    & 0.2209          & 0.2033          & 0.0356          & 0.0090          & 0.0091          & 0.1439          & 0.0758          \\
&FairCFS     & \textbf{0.0117} & \textbf{0.0353} & \textbf{0.0000} & \textbf{0.0000} & \textbf{0.0000} & 0.1356          & \textbf{0.0565}  \\
\hline
\multirow{7}{*}{PE $\downarrow$} &CFS    & 0.1541          & 0.1788          & 0.0141          & 0.0276          & 0.0133          & \textbf{0.0872}          & \textbf{0.0417}          \\
&BAMB    & 0.1433          & 0.1629          & 0.0135          & 0.0260          & 0.0089          & 0.0872          & 0.0454          \\
&HITONMB & 0.2336          & 0.1592          & 0.0129          & 0.0394          & 0.0113          & 0.0872          & 0.0454          \\
&MMMB    & 0.2452          & 0.1600          & 0.0134          & 0.0394          & 0.0112          & 0.0872          & 0.0454          \\
&PCMB    & 0.2099          & 0.1634          & 0.0133          & 0.0394          & 0.0112          & 0.0872          & 0.0454          \\
&STMB    & 0.2114          & 0.1546          & 0.0206          & 0.0370          & 0.0069          & 0.0872          & 0.0454          \\
&FairCFS     & \textbf{0.0204} & \textbf{0.0477} & \textbf{0.0000} & \textbf{0.0000} & \textbf{0.0000} & 0.1444 & 0.0430  \\
\hline
\end{tabular}
\end{table}

\begin{table}[t]
\setlength\tabcolsep{5pt}
\small
\centering
\centering\caption{Comparison of FairCFS, CFS, BAMB, HITONMB, MMMB, PCMB, and STMB on NB Classifier ($\uparrow$ indicates that a higher value of the metric is better, while $\downarrow$ indicates that a lower value of the metric is better).}
\label{tab:my-table}
\begin{tabular}{cc||cccccccc}
\hline
metric &Algorithm        & German          & Compas          & Credit          & Law             & Oulad           & Studentm        & Studentp        \\
\hline
\multirow{7}{*}{ACC $\uparrow$} &CFS    & 0.6950          & 0.6707          & \textbf{0.7797}          & 0.8184          & 0.6783          & 0.9190          & 0.9199          \\
&BAMB    & 0.6970          & 0.6662          & 0.7736          & 0.7966          & 0.6784          & 0.9190          & 0.9168          \\
&HITONMB & \textbf{0.7120} & 0.6707          & 0.7725          & 0.8132          & 0.6699          & 0.9190          & 0.9198          \\
&MMMB    & 0.7070          & 0.6722          & 0.7723          & 0.8106          & 0.6780          & 0.9190          & 0.9167          \\
&PCMB    & 0.6920          & \textbf{0.6748} & 0.7723          & 0.8145          & 0.6780          & 0.9190          & \textbf{0.9214} \\
&STMB    & 0.7010          & 0.6704          & 0.7657          & 0.7799          & 0.6701          & 0.8987          & 0.9198          \\
&FairCFS     & 0.6910          & 0.5790          & 0.7788          & \textbf{0.8550}          & \textbf{0.6786}          & \textbf{0.9191} & 0.9059  \\
\hline
\multirow{7}{*}{SPD $\downarrow$}        &CFS    & 0.1107          & 0.2475          & 0.0357          & 0.0230          & 0.0133          & 0.1566          & 0.1013          \\
&BAMB    & 0.1331          & 0.2208          & 0.0334          & 0.0350          & 0.0132          & 0.1566          & 0.0911          \\
&HITONMB & 0.4403          & 0.2584          & 0.0340          & 0.0338          & 0.0136          & 0.1566          & 0.0949          \\
&MMMB    & 0.4521          & 0.2915          & 0.0360          & 0.0371          & 0.0124          & 0.1566          & 0.0942          \\
&PCMB    & 0.3072          & 0.2783          & 0.0360          & 0.0384          & 0.0124          & 0.1566          & 0.0922          \\
&STMB    & 0.4529          & 0.2204          & 0.0404          & 0.0490          & 0.0116          & 0.1701          & 0.1119          \\
&FairCFS     & \textbf{0.0524} & \textbf{0.0411} & \textbf{0.0000} & \textbf{0.0100} & \textbf{0.0083} & \textbf{0.1423}          & \textbf{0.0844} \\
\hline
\multirow{7}{*}{PE $\downarrow$} &CFS    & 0.1274          & 0.1589          & 0.0236          & 0.0497          & 0.0173          & \textbf{0.0661} & \textbf{0.0565} \\
&BAMB    & 0.1376          & 0.1419          & 0.0217          & 0.0626          & 0.0177          & \textbf{0.0661}          & 0.0837          \\
&HITONMB & 0.3785          & 0.1693          & 0.0225          & 0.0659          & 0.0212          & \textbf{0.0661}          & 0.0837          \\
&MMMB    & 0.3969          & 0.2022          & 0.0246          & 0.0604          & 0.0178          & \textbf{0.0661}          & 0.0794          \\
&PCMB    & 0.3058          & 0.1837          & 0.0246          & 0.0619          & 0.0178          & \textbf{0.0661}          & 0.0770          \\
&STMB    & 0.4272          & 0.1402          & 0.0303          & 0.0791          & 0.0198          & 0.0923          & 0.0819          \\
&FairCFS     & \textbf{0.0524} & \textbf{0.0477} & \textbf{0.0000} & \textbf{0.0241} & \textbf{0.0088} & 0.0734          & 0.0714 \\
\hline
\end{tabular}
\end{table}
In this section, we conducted experiments on three different classifiers: LR, NB, and KNN, and we compared the FairCFS algorithm with CFS, BAMB, HITONMB, MMMB, PCMB, and STMB across seven different datasets. The results, including average accuracy and fairness metrics over 10-fold cross-validation, are summarized in Tables 2, 3, and 4. We can draw the following conclusions:

\begin{table}[h]
\setlength\tabcolsep{5pt}
\small
\centering
\centering\caption{Comparison of FairCFS, CFS, BAMB, HITONMB, MMMB, PCMB, and STMB on KNN Classifier ($\uparrow$ indicates that a higher value of the metric is better, while $\downarrow$ indicates that a lower value of the metric is better).}
\label{tab:my-table}
\begin{tabular}{cc||cccccccc}
\hline
metric &Algorithm        & German          & Compas          & Credit          & Law             & Oulad           & Studentm         &Studentp        \\
\hline
\multirow{7}{*}{ACC $\uparrow$} &CFS    & 0.6520          & \textbf{0.5949} & 0.7145          & 0.8009          & 0.5955          & \textbf{0.9187} & 0.8998          \\
&BAMB    & 0.6550          & 0.5891          & 0.7173          & 0.8078          & 0.5976          & 0.8608          & 0.8952          \\
&HITONMB & 0.6610          & 0.5824          & 0.7171          & 0.8267          & 0.5969          & 0.9112          & 0.8736          \\
&MMMB    & 0.6610          & 0.5792          & \textbf{0.7195}          & 0.8257          & 0.5972          & 0.9112          & 0.8858          \\
&PCMB    & 0.6640          & 0.5780          & 0.7195          & 0.8257          & 0.5972          & 0.9137          & 0.8828          \\
&STMB    & \textbf{0.6780} & 0.5768          & 0.7054          & 0.8278          & 0.5969          & 0.8486          & \textbf{0.9028} \\
&FairCFS     & 0.5880          & 0.5241          & 0.3675          & \textbf{0.8858} & \textbf{0.6786} & 0.8990          & 0.8906  \\
\hline
\multirow{7}{*}{SPD $\downarrow$}        &CFS    & 0.1475          & 0.1605          & 0.0162          & 0.0144          & 0.0160          & 0.1161          & 0.0747          \\
&BAMB    & 0.1841          & 0.1744          & 0.0176          & 0.0154          & 0.0290          & 0.1106          & 0.0786          \\
&HITONMB & 0.1569          & 0.1702          & 0.0176          & 0.0291          & 0.0274          & 0.1111          & 0.0699          \\
&MMMB    & 0.1569          & 0.1625          & 0.0226          & 0.0292          & 0.0294          & 0.1111          & 0.0907          \\
&PCMB    & 0.2759          & 0.1664          & 0.0225          & 0.0292          & 0.0294          & 0.1261          & 0.1155          \\
&STMB    & 0.0901          & 0.1367          & 0.0264          & 0.0319          & 0.0270          & \textbf{0.1042}          & \textbf{0.0464} \\
&FairCFS     & \textbf{0.0754} & \textbf{0.0197} & \textbf{0.0101} & \textbf{0.0013} & \textbf{0.0111} & 0.1130          & 0.0835 \\
\hline
\multirow{7}{*}{PE $\downarrow$} &CFS    & 0.1811          & 0.1513          & 0.0171          & 0.0391          & 0.0213          & 0.0712          & 0.0635          \\
&BAMB    & 0.2091          & 0.1352          & 0.0168          & 0.0389          & 0.0432          & 0.1487          & 0.0686          \\
&HITONMB & 0.1569          & 0.1436          & 0.0170          & 0.0665          & 0.0332          & 0.0912          & 0.0653          \\
&MMMB    & 0.1593          & 0.1171          & 0.0179          & 0.0645          & 0.0412          & 0.0912          & 0.0653          \\
&PCMB    & 0.2705          & 0.1375          & 0.0177          & 0.0645          & 0.0412          & 0.0712          & 0.0639          \\
&STMB    & 0.1000          & 0.1191          & 0.0326          & 0.0686          & 0.0353          & 0.1341          & \textbf{0.0512} \\
&FairCFS     & \textbf{0.0717} & \textbf{0.0488} & \textbf{0.0118}  & \textbf{0.0139} & \textbf{0.0103} & \textbf{0.0603}          & 0.0672 \\
\hline
\end{tabular}
\end{table}

\textbf{Accuracy: }Next, we introduce the accuracy comparison results of FairCFS with others.

\begin{itemize}

\item \textbf{LR Classifier}: In Table 2, it can be observed that, compared to other causal feature selection algorithms, the FairCFS algorithm maintains a similar level of accuracy on most datasets, with an average difference of around 0.01. CFS, HITONMB, and MMMB all achieved the best accuracy on both datasets, while neither PCMB nor STMB achieved the best accuracy on either dataset. However, from each data set, the difference in algorithm accuracy is minimal, and the average accuracy variance in the causal feature selection algorithm does not exceed 0.01. This is because, according to the faithfulness assumption, the MB selected by the causal feature selection algorithm should be unique. When the MB discovery algorithm recognizes MB correctly, The MB sets discovered by different algorithms are also different. The FairCFS algorithm removes unfair nodes from MB, and the accuracy will inevitably decrease. Still, it can be seen from the experimental results that the accuracy of the FairCFS algorithm does not decline much, which is comparable to other algorithms in the Logistic Regression Classifier.

\item \textbf{NB Classifier}: In Table 3, for the NB classifier, FairCFS not only achieved the highest accuracy on the Studentm dataset but also on the Law and Oulad dataset. On the Law dataset, FairCFS is about 0.04 percent higher than the second place. PCMB had the best accuracy on two datasets, and CFS and HITONMB each had one dataset with the best accuracy. However, except for the Compas dataset, the difference between FairCFS and the best results on other datasets is less than 0.02. It can be seen that for the Naive Bayes Classifier, FairCFS performed slightly better but did not produce a significant difference. It can be stated that according to hypothesis 1 mentioned in the FairCFS algorithm, the distribution of data after intervening in the causal plot is more in line with the Naive Bayes classifier.

\item \textbf{KNN Classifier}: In Table 4, on the KNN classifier, FairCFS, CFS, and STMB still achieved the highest accuracy on two datasets, and FairCFS surpassed the second-best algorithm by 0.06 on the Law dataset. It can be seen that there is a slightly larger difference with other algorithms on the dataset where FairCFS achieves the best accuracy. There is a difference of 0.06 from the second place on the Law dataset and 0.08 on the Oulad dataset from the second place. However, FAirCFS also had a dataset called Credit that significantly reduces the accuracy rate because the Credit dataset had more features that have a direct impact on the gender of sensitive attributes. More features were deleted in FairCFS, and important prediction information is lost.
\end{itemize}

\textbf{Fairness}: Based on the remaining two metrics' results from Tables 2, 3, and 4, we can draw the following conclusions:
\begin{itemize}
\item \textbf{LR Classifier}: Table 2 shows that for the LR classifier, FairCFS achieved the best fairness results on six datasets. Notably, it significantly improved fairness on datasets like German and Compas compared to other causal feature selection algorithms. While FairCFS does not achieve the best fairness on the Studentm dataset, the difference in fairness metrics is slight, around 0.01, compared to the best result. However, For the indicator PE, FairCFS and the best effect have a large gap, about 0.06 or so, through the analysis of the data set, maybe because the number of samples in this data set is too small, resulting in features that can not distinguish the difference of samples, from these causal feature selection algorithms also can be seen that the indicators on this dataset are almost the same.

\item \textbf{NB Classifier}: In Table 3, for the NB classifier, FairCFS achieved the best fairness results on all datasets for SPD. While CFS performed better on the last two datasets for PE. Among them, on the dataset Compas, except for the FairCFS indicator, SPD is lower than 0.05, and the SPD value of other algorithms is higher than 0.22; this result can reflect that the MB of the Compas dataset class variable contains a large amount of sensitive information, FairCFS can significantly reduce unfairness by finding unfair features to delete it. However, the improvement of fairness also led to a decrease in the accuracy of FairCFS on Compas, verifying the inherent trade-off between accuracy and fairness.

\item \textbf{KNN Classifier}: In Table  4, FairCFS continues to demonstrate its advantages in fairness on six datasets. On the first five datasets, FairCFS performed best on both fairness indicators and still reduced both fairness indicators by about 0.10 on the Compas dataset. On the Studentp dataset, the fairness achieved by the FairCFS algorithm is also not significantly different from the best-performing approach. However, it also highlights FairCFS's limitations on small-scale datasets, where the reliance on conditional independence tests may affect its performance.
\end{itemize}

\begin{figure}[t]
    \centering
    \vspace{1.95em}
    \begin{subfigure}[t]{0.5\linewidth}
        \includegraphics[width=\linewidth]{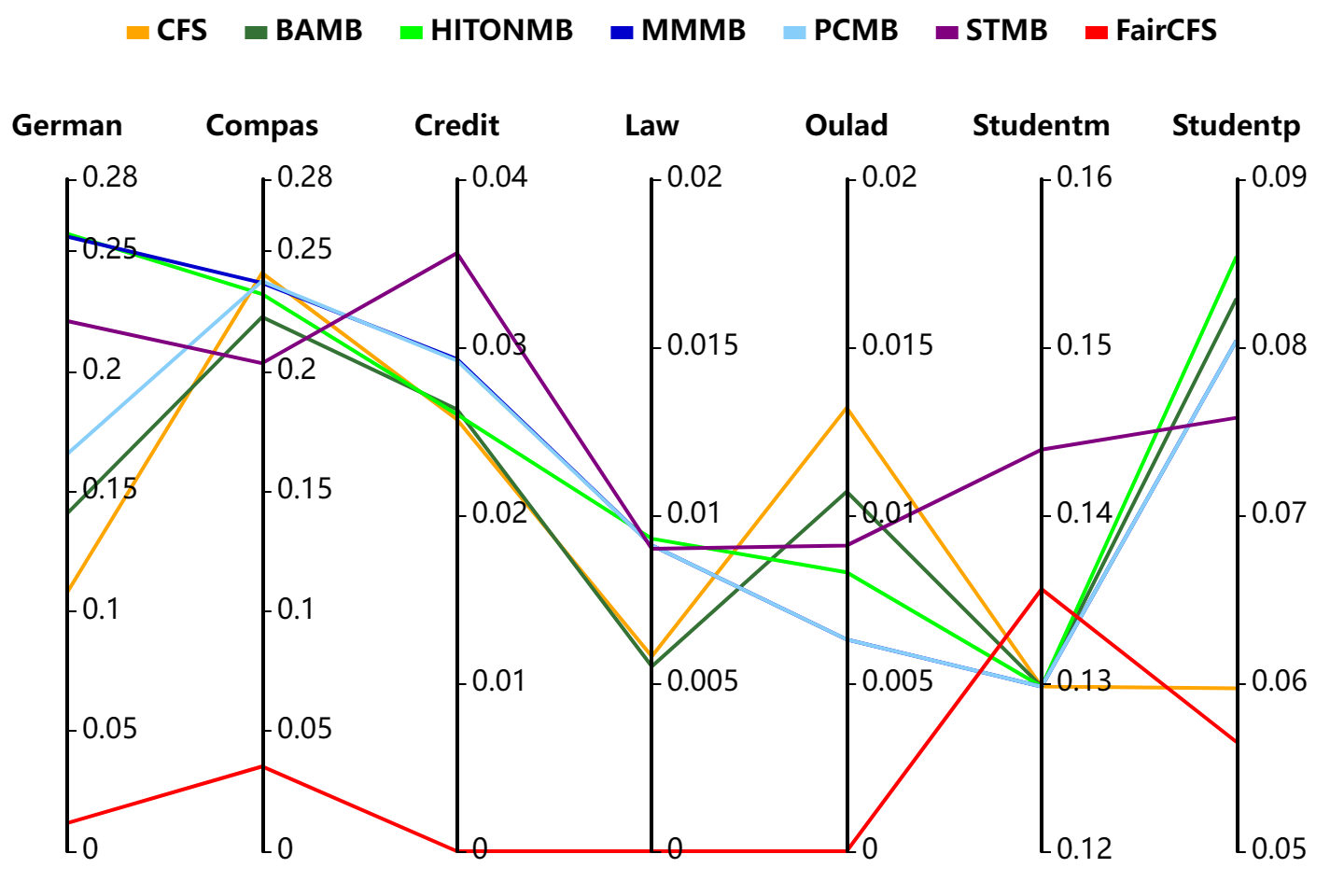}
    \end{subfigure}
    \hspace{-0.7em}
    \begin{subfigure}[t]{0.5\linewidth}
        \includegraphics[width=\linewidth]{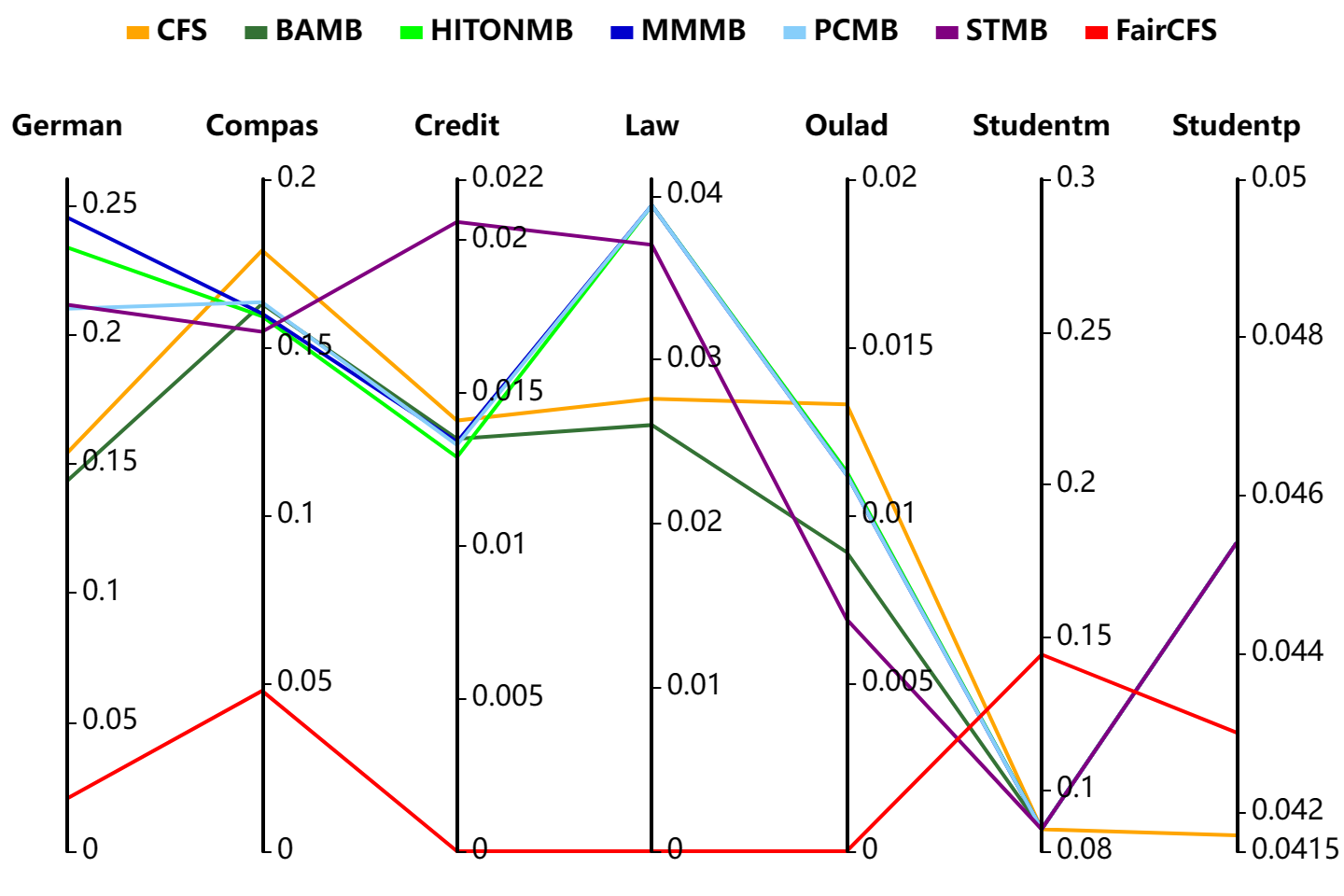}
    \end{subfigure}
    %\vspace{0.001em}
    \caption{Line chart of FairCFS and its causal feature selection rivals on fairness metrics SPD (left) and PE (right) with LR classifier.}
\end{figure}

\begin{figure}[t]
    \centering
    \begin{subfigure}[t]{0.5\linewidth}
        \includegraphics[width=\linewidth]{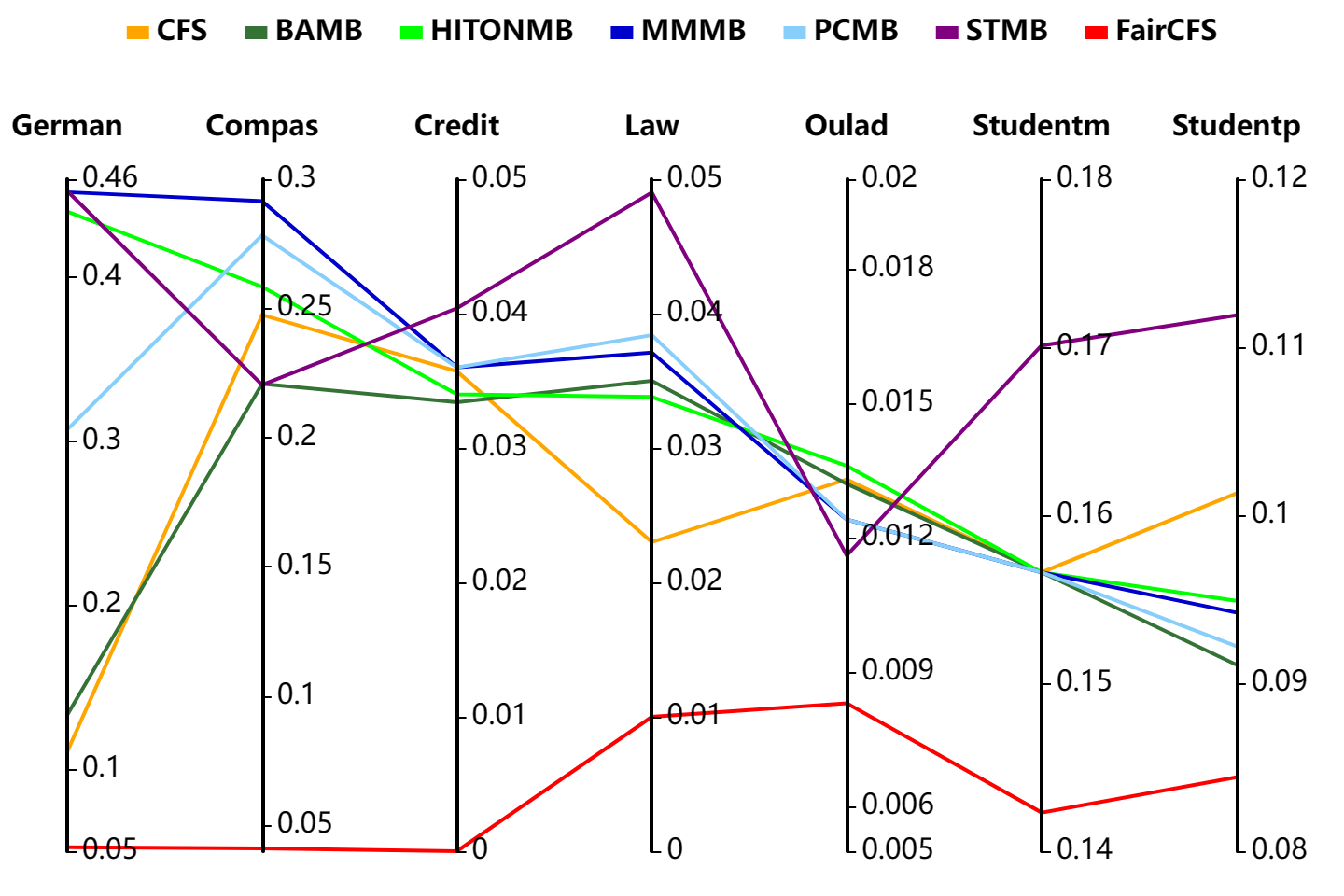}
    \end{subfigure}
    \hspace{-0.7em}
    \begin{subfigure}[t]{0.5\linewidth}
        \includegraphics[width=\linewidth]{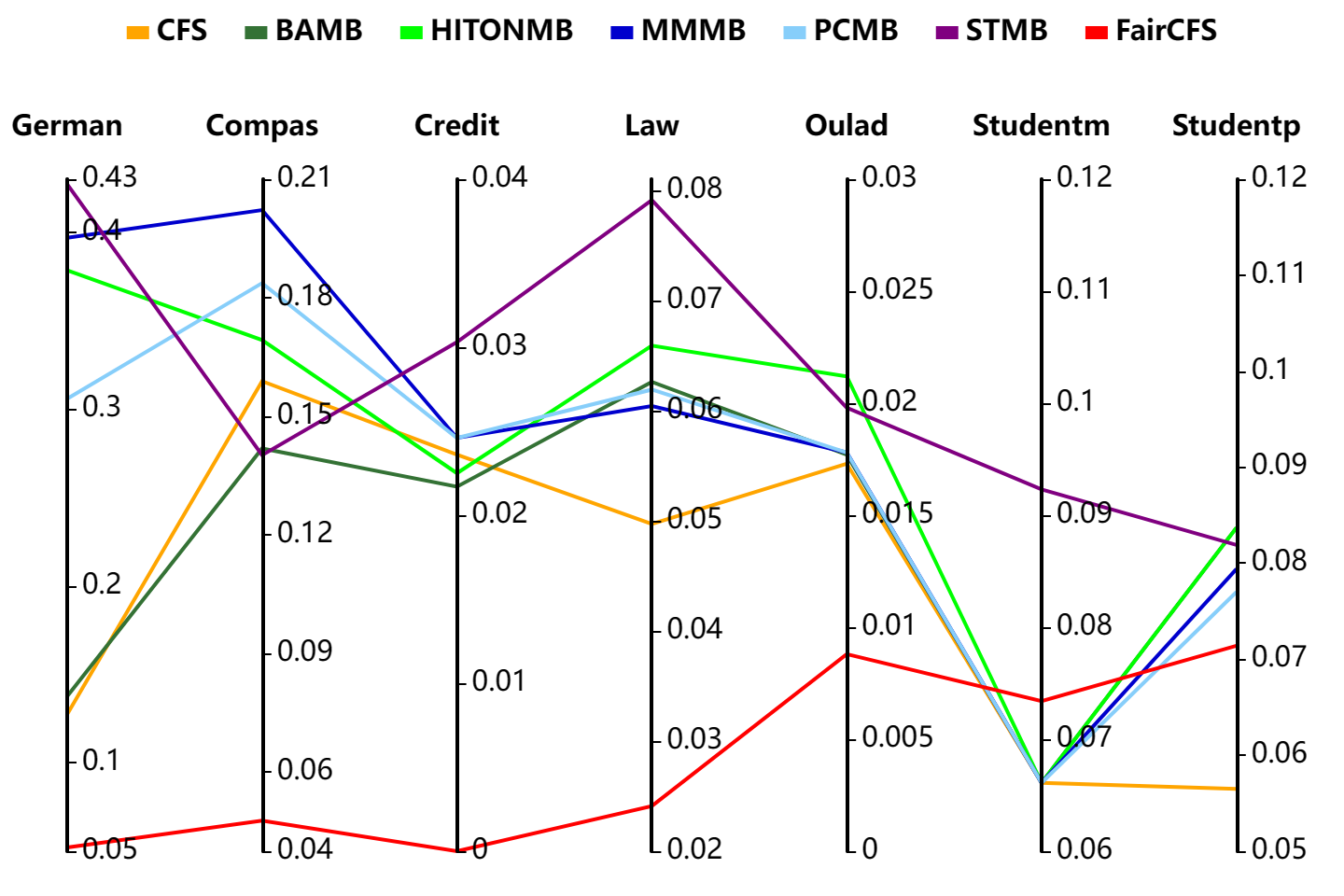}
    \end{subfigure}
    %\vspace{0.001em}
    \caption{Line chart of FairCFS and its causal feature selection rivals on fairness metrics SPD (left) and PE (right) with NB classifier.}
\end{figure}

\begin{figure}[t]
    \centering
    \begin{subfigure}[t]{0.5\linewidth}
        \includegraphics[width=\linewidth]{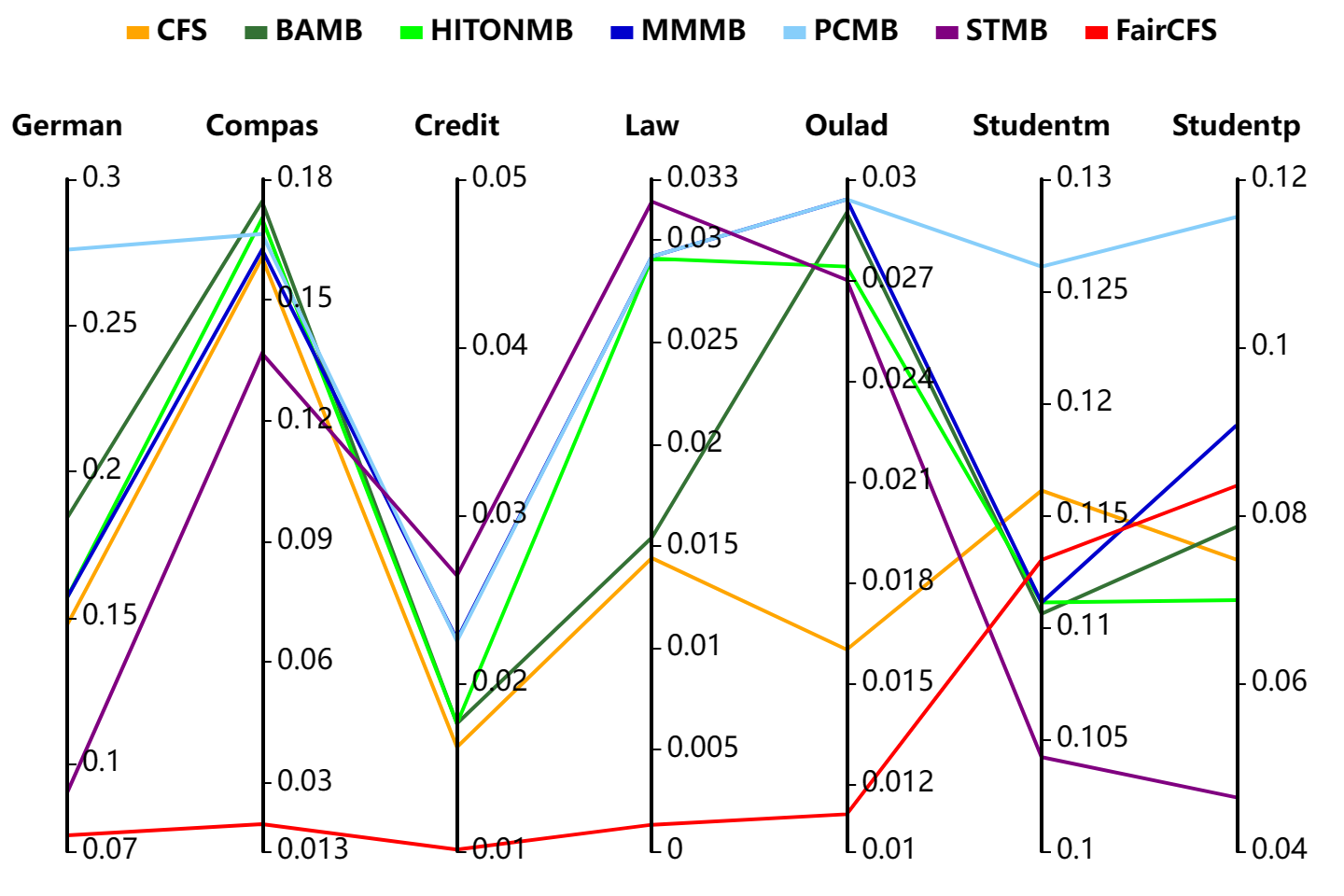}
    \end{subfigure}
    \hspace{-0.7em}
    \begin{subfigure}[t]{0.5\linewidth}
        \includegraphics[width=\linewidth]{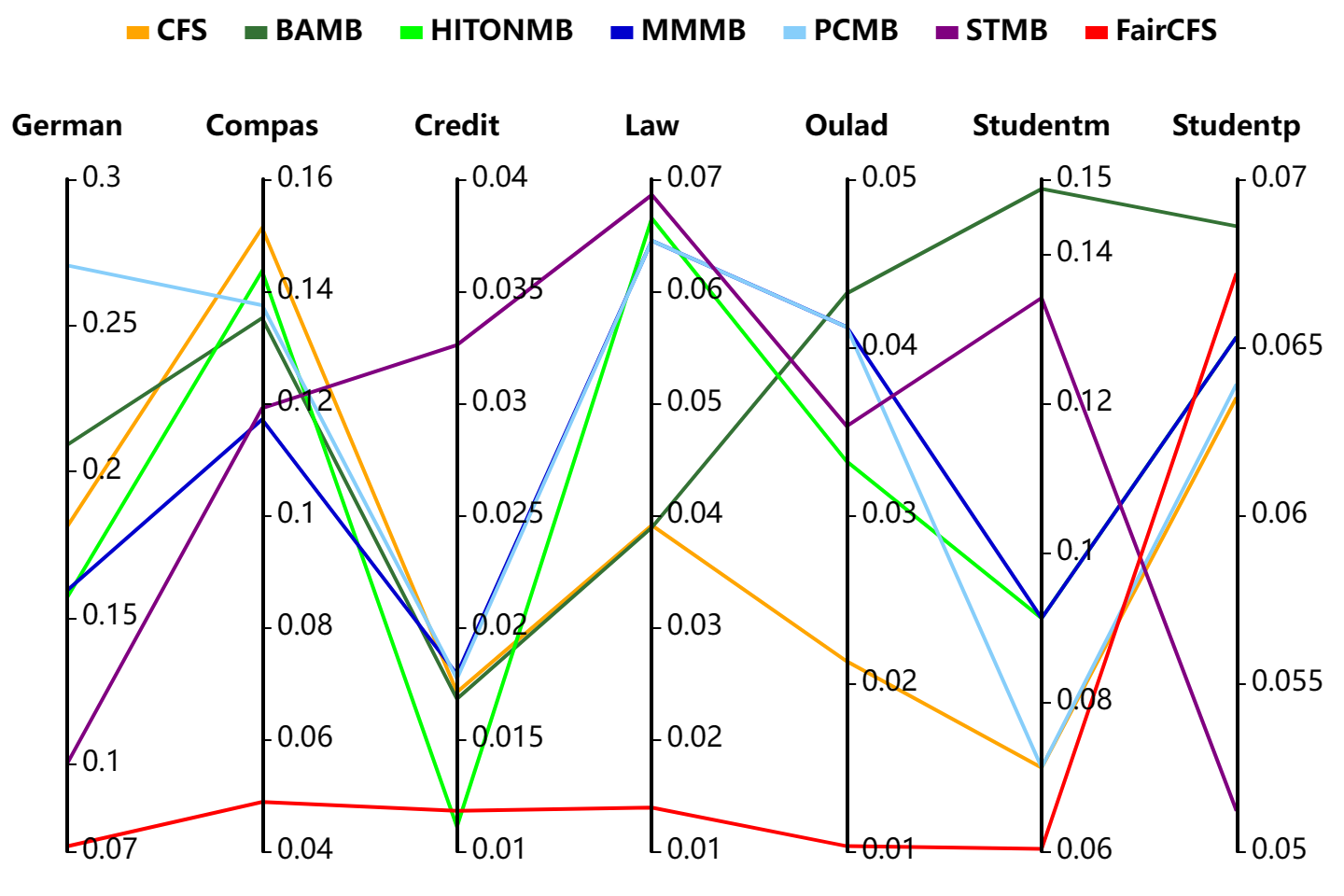}
    \end{subfigure}
    %\vspace{0.001em}
    \caption{Line chart of FairCFS and its causal feature selection rivals on fairness metrics SPD (left) and PE (right) with KNN classifier.}

\end{figure}
To visually highlight the FairCFS algorithm's superiority in fairness compared to six causal feature selection algorithms, we provided comparative line charts for FairCFS and the other algorithms in Figure 3. In terms of fairness, Figs. 3, 4, and 5 show that FairCFS contains the lowest fairness indicator on most datasets, especially the first five, where FairCFS is at the lowest point. This shows that when FairCFS looks for fair causal features, it effectively removes unfair features from MB. On the small sample dataset, the gap between the comparison algorithms cannot be opened because when looking for accurate MB, the causal feature selection algorithm needs enough sample size to obtain a reliable conditional independence test, and when the sample size is not enough, it may not be able to accurately identify the causal relationship between features.

\subsection{Comparison of FairCFS with fair feature selection algorithm}
\begin{table}[h]
\setlength\tabcolsep{5pt}
\small
\centering
\centering\caption{Comparison of FairCFS, Auto, Seqsel on LR Classifier ($\uparrow$ indicates that a higher value of the metric is better, while $\downarrow$ indicates that a lower value of the metric is better).}
\label{tab:my-table}
\begin{tabular}{cc||cccccccc}
\hline
metric &Algorithm        & German & Compas & Credit & Law    & Oulad  & Studentm & Studentp \\
\hline
\multirow{3}{*}{ACC $\uparrow$} &Auto    & \textbf{0.6990} & \textbf{0.6791} & \textbf{0.7969} & \textbf{0.8897} & \textbf{0.6866} & 0.8129   & 0.8689   \\
&Seqsel  & 0.6980 & 0.6296 & 0.7788 & 0.8896 & 0.6808 & 0.8205   & 0.8752   \\
&FairCFS     & 0.6980 & 0.5722 & 0.7788 & \textbf{0.8897} & 0.6796 & \textbf{0.9189}   & \textbf{0.9275}    \\
\hline
\multirow{3}{*}{SPD $\downarrow$}        &Auto    & 0.1156          & 0.2418          & 0.0314          & \textbf{0.0000}          & 0.0194          & 0.1818          & 0.0907          \\
&Seqsel  & 0.0718          & 0.1044          & 0.0000          & 0.0005          & 0.0090          & \textbf{0.0951}          & 0.0935          \\
&FairCFS     & \textbf{0.0117} & \textbf{0.0353} & \textbf{0.0000} & \textbf{0.0000} & \textbf{0.0000} & 0.1356          & \textbf{0.0565}  \\
\hline
\multirow{3}{*}{PE $\downarrow$} &Auto    & 0.1325          & 0.1702          & 0.0212          & \textbf{0.0000}          & 0.0182          & 0.2965          & 0.0584          \\
&Seqsel  & 0.0569          & 0.0671          & \textbf{0.0000}          & 0.0017          & 0.0099          & 0.3095          & \textbf{0.0381} \\
&FairCFS     & \textbf{0.0204} & \textbf{0.0477} & \textbf{0.0000} & \textbf{0.0000} & \textbf{0.0000} & \textbf{0.1444} & 0.0430  \\
\hline
\end{tabular}
\end{table}

\begin{table}[]
\setlength\tabcolsep{5pt}
\small
\centering
\centering\caption{Comparison of FairCFS, Auto, Seqsel on NB Classifier.}
\label{tab:my-table}
\begin{tabular}{cc||cccccccc}
\hline
metric &Algorithm        & German          & Compas          & Credit          & Law             & Oulad           & Studentm        & Studentp        \\
\hline
\multirow{3}{*}{ACC $\uparrow$} &Auto    & 0.6840          & \textbf{0.6800}          & \textbf{0.7970} & 0.6775          & \textbf{0.6790} & 0.7797          & 0.8721          \\
&Seqsel  & 0.6210          & 0.6296          & 0.2935          & \textbf{0.8555}          & 0.6786          & 0.7925          & 0.8582          \\
&FairCFS     & \textbf{0.6910}          & 0.5790          & 0.7788          & 0.8550          & 0.6786          & \textbf{0.9190} & \textbf{0.9059}  \\
\hline
\multirow{3}{*}{SPD $\downarrow$}        &Auto    & 0.1450          & 0.2997          & 0.0282          & 0.0326          & 0.0124          & \textbf{0.0924}          & 0.1666          \\
&Seqsel  & 0.1187          & 0.1051          & 0.0136          & 0.0125          & 0.0129          & 0.1176          & 0.1222          \\
&FairCFS     & \textbf{0.0524} & \textbf{0.0411} & \textbf{0.0000} & \textbf{0.0100} & \textbf{0.0083} & 0.1423          & \textbf{0.0844} \\
\hline
\multirow{3}{*}{PE $\downarrow$} &Auto    & 0.1656          & 0.2036          & 0.0165          & 0.0480          & 0.0137          & 0.2000          & 0.1124          \\
&Seqsel  & 0.0875          & 0.0666          & 0.0191          & 0.0552          & 0.0122          & 0.2121          & 0.0941          \\
&FairCFS     & \textbf{0.0521} & \textbf{0.0477} & \textbf{0.0000} & \textbf{0.0241} & \textbf{0.0088} & \textbf{0.0734}          & \textbf{0.0714} \\
\hline
\end{tabular}
\end{table}

\begin{table}[]
\setlength\tabcolsep{5pt}
\small
\centering
\centering\caption{Comparison of FairCFS, Auto, Seqsel on KNN Classifier.}
\label{tab:my-table}
\begin{tabular}{cc||cccccccc}
\hline
metric &Algorithm        & German          & Compas          & Credit          & Law             & Oulad           & Studentm        & Studentp        \\
\hline
\multirow{3}{*}{ACC $\uparrow$} &Auto    & \textbf{0.6660}          & \textbf{0.5928}          & \textbf{0.7884} & 0.7800          & 0.6600          & 0.6987          & 0.8243          \\
&Seqsel  & 0.6470          & 0.5904          & 0.4374          & 0.7093          & 0.6379          & 0.6885          & 0.8351          \\
&FairCFS     & 0.5880          & 0.5241          & 0.3675          & \textbf{0.8858} & \textbf{0.6786} & \textbf{0.8990}          & \textbf{0.8906}  \\
\hline
\multirow{3}{*}{SPD $\downarrow$}        &Auto    & 0.0855          & 0.1941          & 0.0507          & 0.0255          & 0.0162          & 0.1470          & 0.0910          \\
&Seqsel  & 0.0999          & 0.1100          & 0.0166          & 0.0198          & 0.0120          & \textbf{0.0992} & \textbf{0.0774}          \\
&FairCFS     & \textbf{0.0754} & \textbf{0.0197} & \textbf{0.0101} & \textbf{0.0013} & \textbf{0.0111} & 0.1130          & 0.0835 \\
\hline
\multirow{3}{*}{PE $\downarrow$} &Auto    & 0.1162          & 0.1690          & 0.0235          & 0.0726          & 0.0164          & 0.2047          & 0.0961          \\
&Seqsel  & 0.1507          & 0.0936          & 0.0177          & 0.0815          & 0.0130          & 0.3337          & \textbf{0.0622}          \\
&FairCFS     & \textbf{0.0717} & \textbf{0.0488} & \textbf{0.0118}  & \textbf{0.0139} & \textbf{0.0103} & \textbf{0.0603}          & 0.0672 \\
\hline
\end{tabular}
\end{table}
In this section, we compared the FairCFS algorithm with Auto and Seqsel across seven different datasets. The results, including average accuracy and fairness metrics over 10-fold cross-validation, are summarized in Tables 5, 6, and 7. We can draw the following conclusions:

\begin{figure}[t]
    \centering
    \vspace{1.98em}
    \begin{subfigure}[t]{0.5\linewidth}
        \includegraphics[width=\linewidth]{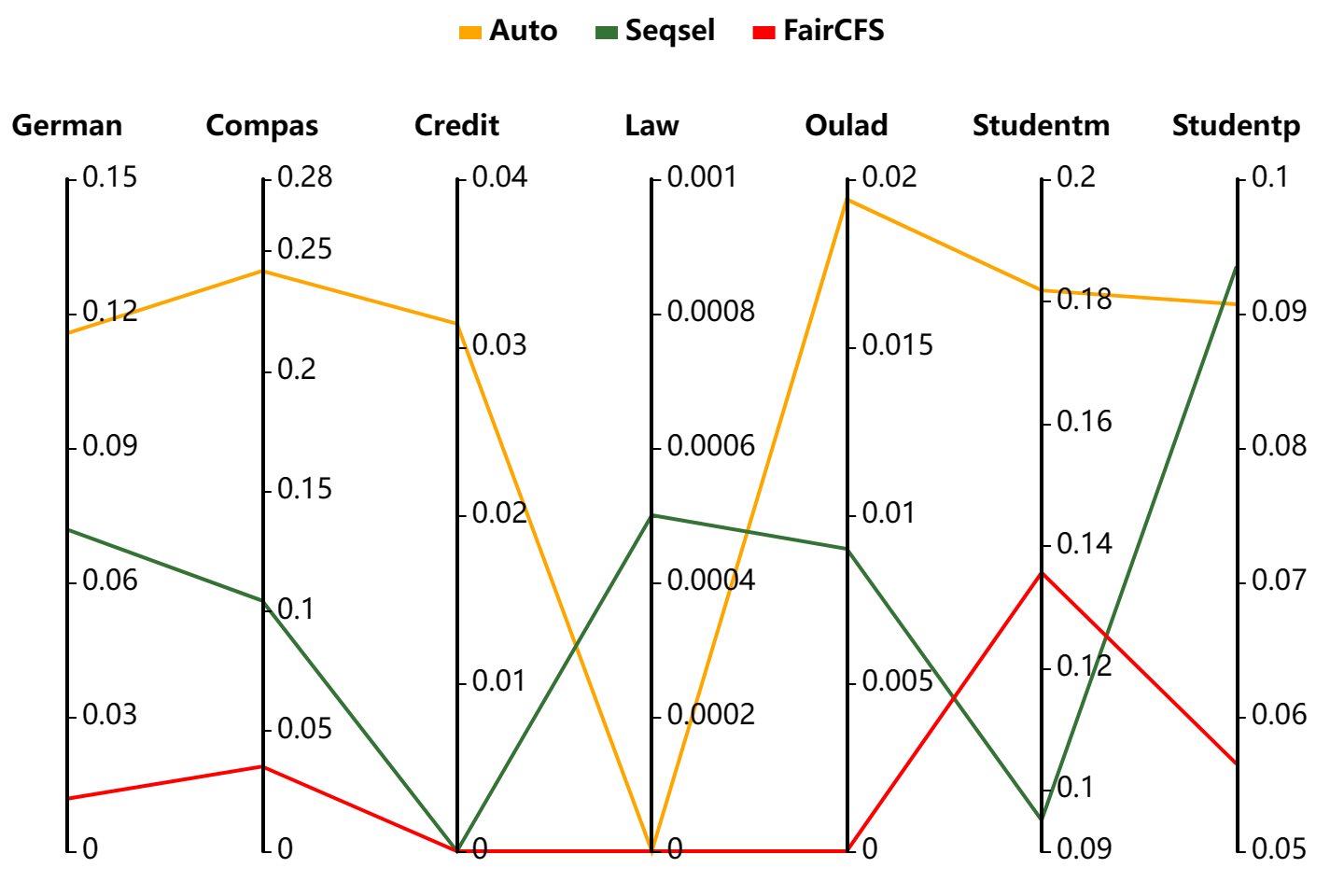}
    \end{subfigure}
    \hspace{-0.7em}
    \begin{subfigure}[t]{0.5\linewidth}
        \includegraphics[width=\linewidth]{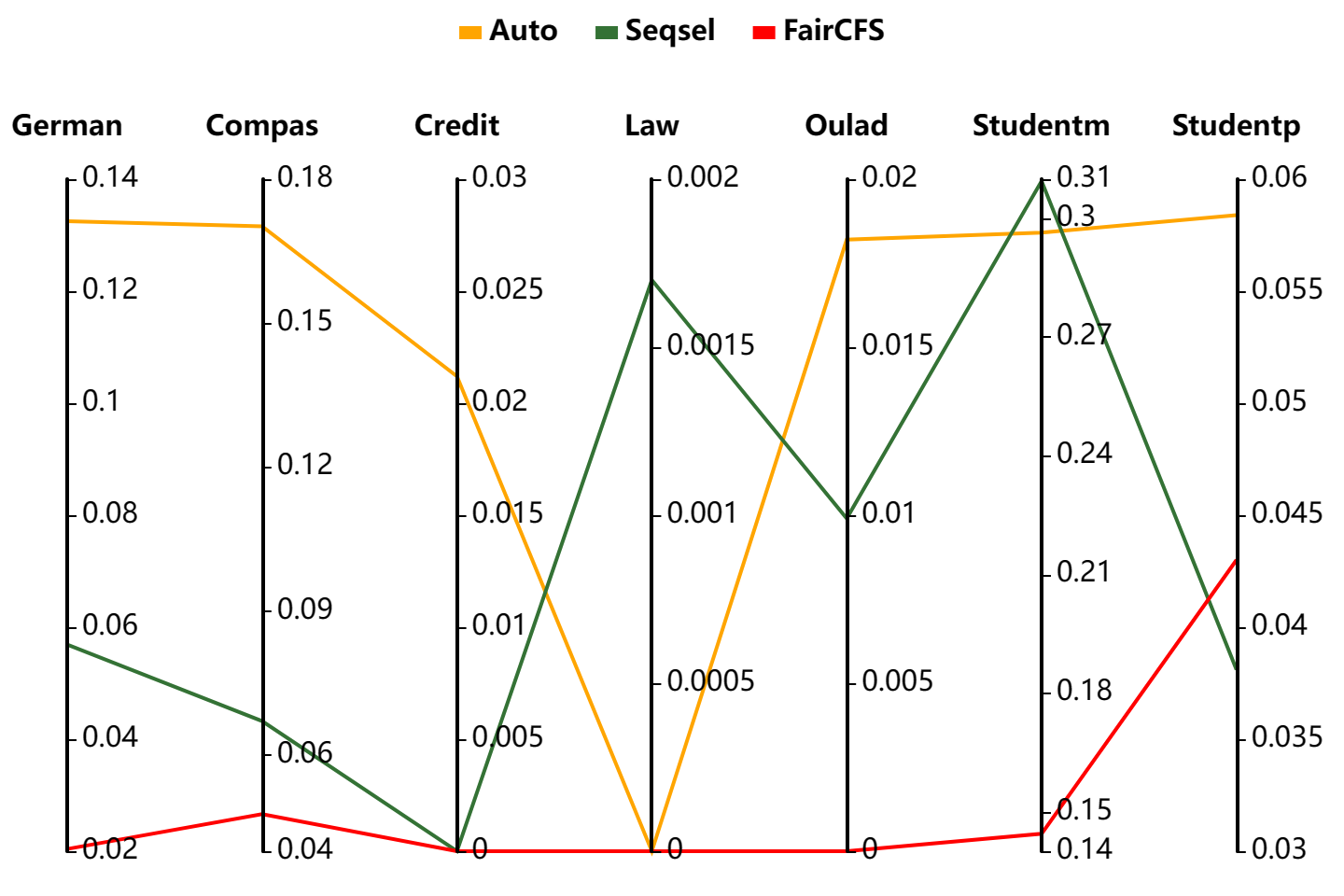}
    \end{subfigure}
    \caption{Line chart of FairCFS and its fair feature selection rivals on fairness metrics SPD (left) and PE (right) with LR classifier.}
\end{figure}

\begin{figure}[t]
    \centering
    \begin{subfigure}[t]{0.5\linewidth}
        \includegraphics[width=\linewidth]{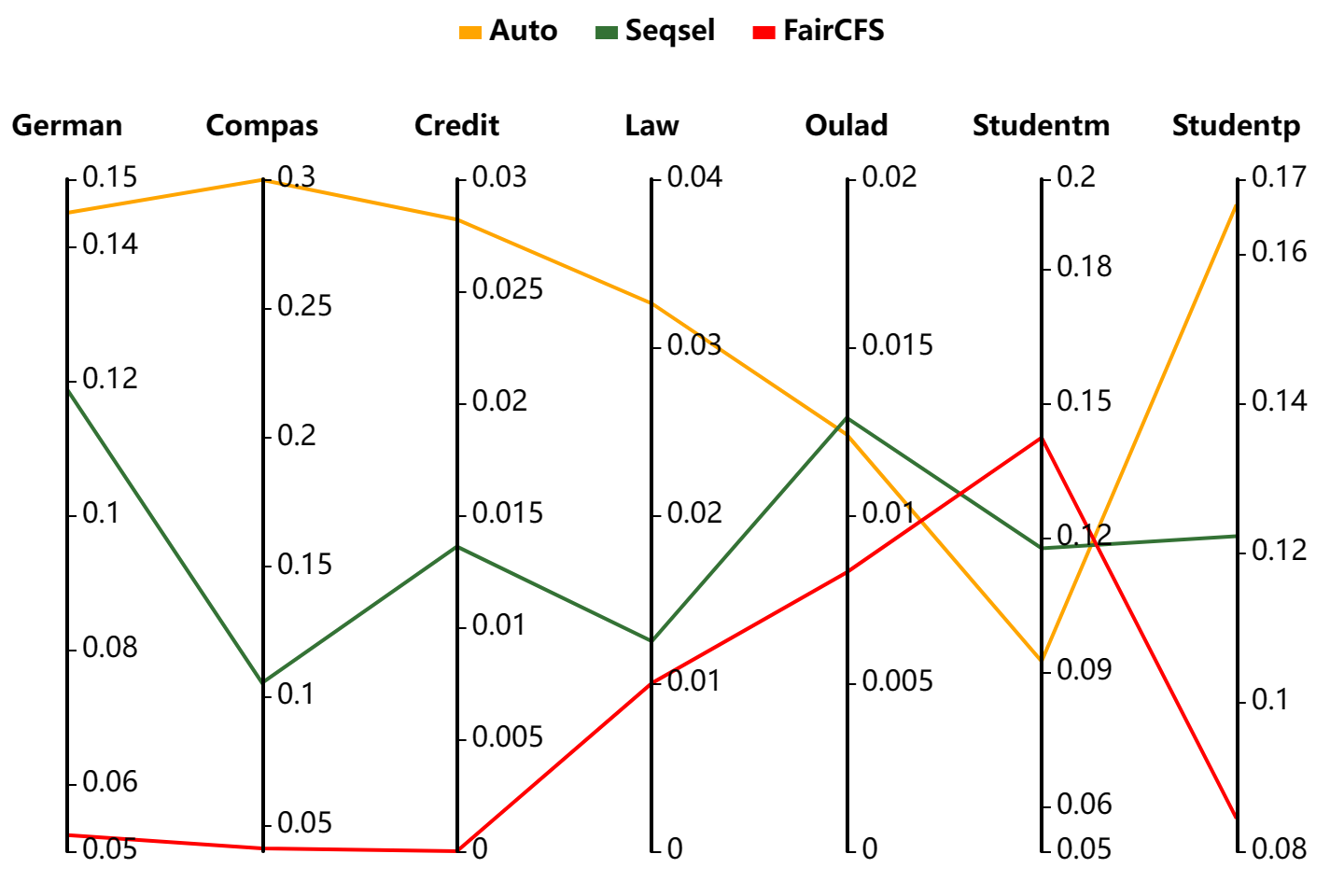}
    \end{subfigure}
    \hspace{-0.7em}
    \begin{subfigure}[t]{0.5\linewidth}
        \includegraphics[width=\linewidth]{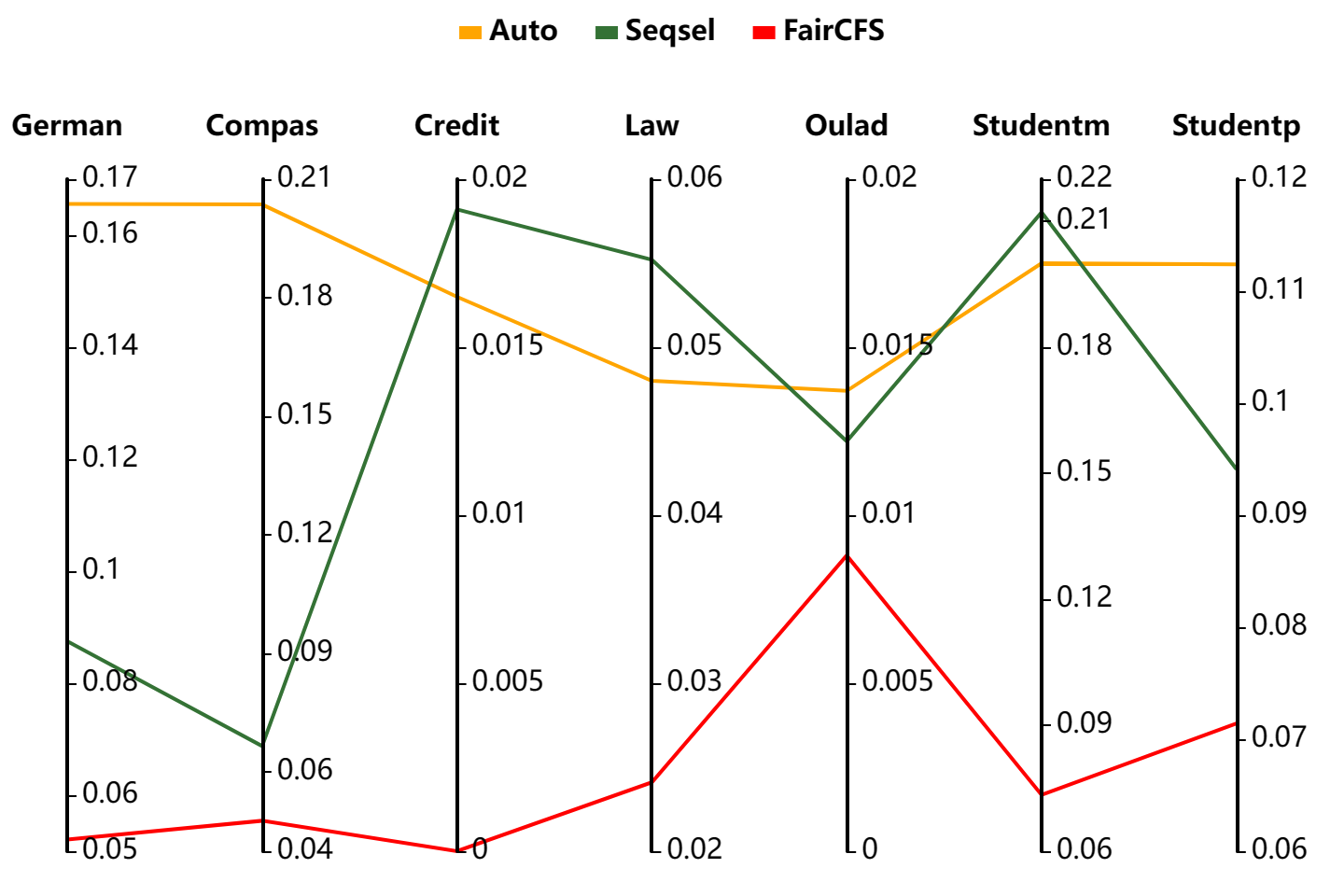}
    \end{subfigure}
    \caption{Line chart of FairCFS and its fair feature selection rivals on fairness metrics SPD (left) and PE (right) with NB classifier.}
\end{figure}

\begin{figure}[t]
    \centering
    \begin{subfigure}[t]{0.5\linewidth}
        \includegraphics[width=\linewidth]{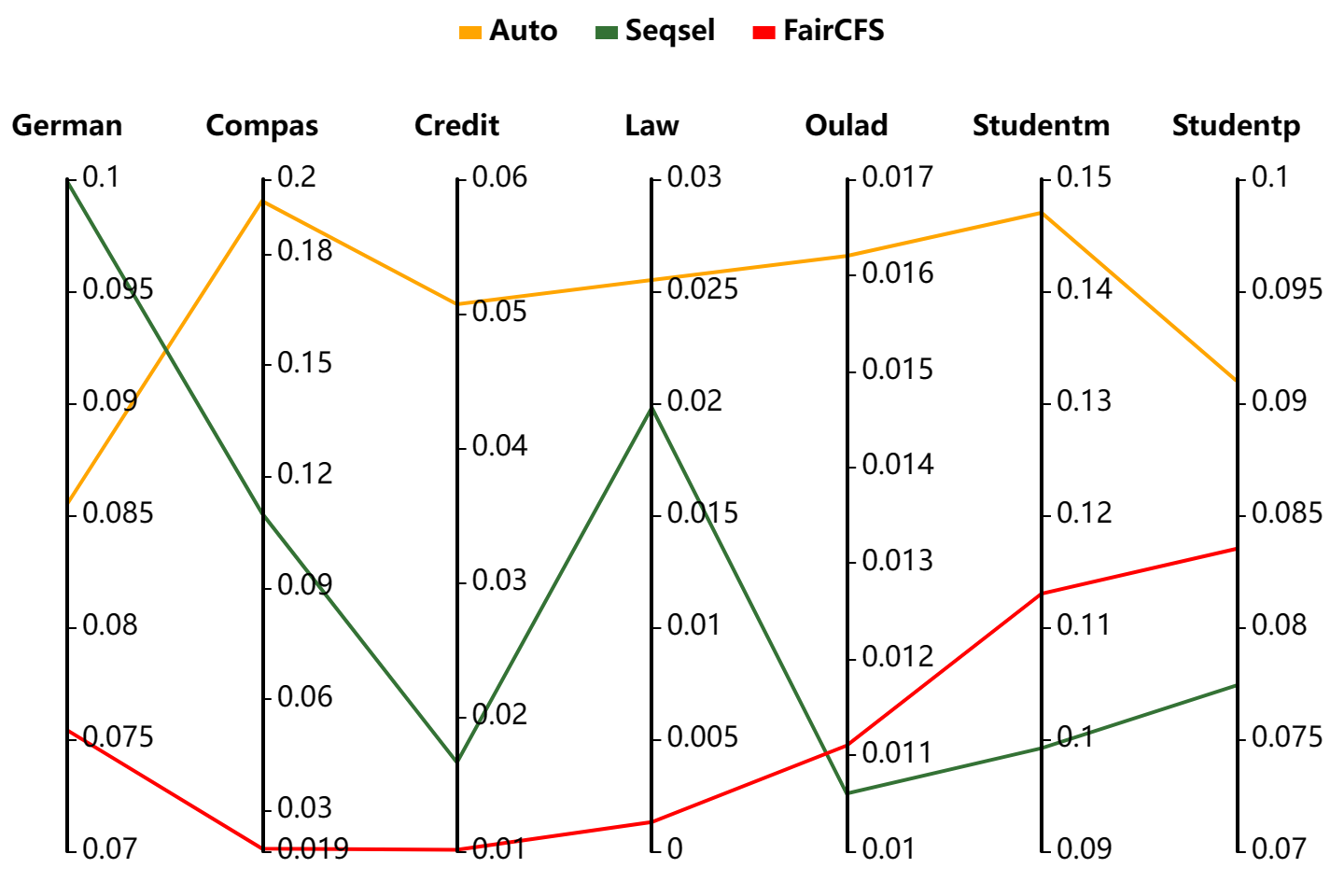}
    \end{subfigure}
    \hspace{-0.7em}
    \begin{subfigure}[t]{0.5\linewidth}
        \includegraphics[width=\linewidth]{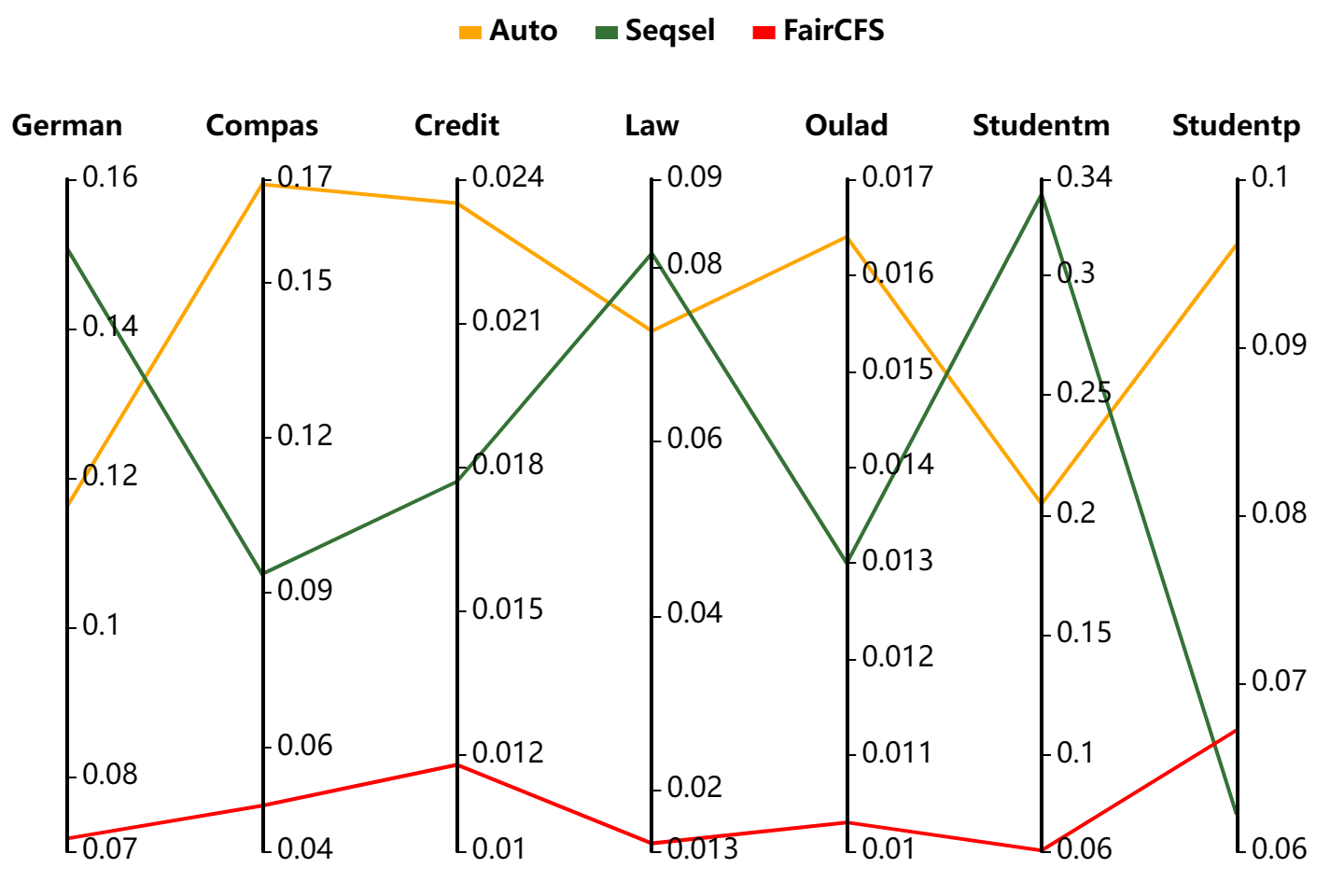}
    \end{subfigure}
    \caption{Line chart of FairCFS and its fair feature selection rivals on fairness metrics SPD (left) and PE (right) with KNN classifier.}
\end{figure}

\textbf{Accuracy}: The performance of FairCFS and Auto algorithms in terms of the Accuracy metric in Tables 5, 6, and 7 indicates that both FairCFS and Auto outperform other algorithms. They achieved the best accuracy on 3 to 4 different datasets. On the other hand, the Seqsel algorithm only achieved the best accuracy on one dataset using NB classifiers. It can be seen the performance of these algorithms can vary depending on their feature selection strategies. FairCFS and Auto performed well by considering both predictive accuracy and fairness, while Seqsel's focus on fairness alone can lead to lower accuracy in some cases. When selecting features, Auto's strategy selects features with more accuracy because FairCFS relies on the conditional independence relationship between features and class variables and sensitive variables when selecting features, and some statistic-based practices may find more features than FairCFS, thus containing more prediction information and higher accuracy.

\textbf{Fairness}: Tables 5, 6, and 7 clearly illustrated that the FairCFS algorithm can achieve higher fairness while maintaining good accuracy. To visually highlight the FairCFS algorithm's superiority in fairness compared to two fair feature selection algorithms, we provided comparative line charts for FairCFS and the other algorithms in Figure 4. In terms of fairness, Figs. 6, 7, and 8 show that FairCFS contained the lowest fairness indicator on most datasets, especially the first five, where FairCFS is at the lowest point.

FairCFS can achieve the best fairness on the first five datasets and three classifiers, among which the fairness index can be reduced to 0 on the dataset Credit and Law. These data sets are all large sample datasets, which makes the FairCFS conditional independent test effect good, and it is easier to find the fair features between the datasets. At the same time, the Auto algorithm is limited by the indicators when the model is trained, resulting in the inability to find the features related to other fairness indicators clearly, and the model fairness is the worst. Seqsel algorithm finds features when the judgment of fair features is limited by acceptable features; in large-scale datasets, admissible features are difficult to specify, so the judgment of fair features is prone to errors so that the best fairness is not achieved. This demonstrates that by searching for the Markov blankets of sensitive variables to find a conditional set that makes features independent of the sensitive variable, it is possible to effectively block the transmission of sensitive information.

However, an analysis of datasets where FairCFS performed less favorably in terms of fairness reveals that on small-scale datasets, FairCFS may not always exhibit the best fairness performance. This is because when the number of data samples is small, the reliability of conditional independence tests can be challenging, and they may not fully discover the causal relationships between features, leading to errors in the identification of fair features.

\section{Conclusion}
This paper analyzed the issue of blocking the transmission of sensitive information in fair causal feature selection based on intervention fairness. Subsequently, we proposed a fair causal feature selection algorithm, FairCFS, which identifies features independent of sensitive variables by discovering the Markov blankets of both the class and sensitive variables. Finally, experiments on seven real-world datasets demonstrated that FairCFS achieves comparable accuracy and better fairness. However, it is worth noting that the conditional independence tests based on the $G^{2}$ test used in FairCFS may not be sufficient when the sample size of the dataset is small, leading to unexpected results. Therefore, future research could focus on reliably improving fairness when dealing with small sample datasets.

\begin{acks}

This work was supported by the National Key Research and Development Program of China (under grant 2020AAA0106100), the National Natural Science Foundation of China (under grant 62306002, 62176001, 62376087, and U1936220), and the Natural Science Project of Anhui Provincial Education Department (under grant 2023AH030004).

\end{acks}

%%
%% The next two lines define the bibliography style to be used, and
%% the bibliography file.
\bibliographystyle{ACM-Reference-Format}
\bibliography{References}

\end{document}